\documentclass[11pt]{article}
\usepackage[utf8]{inputenc}
\pdfoutput=1
\usepackage{setspace}
\usepackage{palatino}
\usepackage{graphicx}
\usepackage{float}
\usepackage{titling} 
\usepackage{multirow}
\usepackage{lscape}
\usepackage{amsmath}
\usepackage{amssymb}
\usepackage[a4paper, total={6in, 9.5in}]{geometry}
\usepackage{array}
\usepackage[normalem]{ulem}
\fontfamily{ppl}\selectfont 
\usepackage{eqparbox}
\usepackage{arydshln}
\usepackage{mathrsfs}
\usepackage[american]{babel}
\usepackage{mathtools}

\usepackage{rotating}
\usepackage{amsthm}
\newtheorem{theorem}{Theorem}

\newtheorem{definition}{Definition}

\usepackage[ruled,vlined]{algorithm2e}

\usepackage{bm}
\usepackage{caption}
\usepackage{subcaption}

\newcommand{\appendixqqsection}[1]{\addtocounter{section}{1}
   \setcounter{table}{0}
   \setcounter{figure}{0}
   \setcounter{equation}{0}
   \setcounter{subsection}{0}
  \section*{Supplementary \Alph{section}: #1}
}

\newcommand\appendixqq{
   \setcounter{section}{0}
   \renewcommand\thesection{\Alph{section}}
   \renewcommand{\thesubsection}{\thesection. \arabic{subsection}}
   \renewcommand{\thesubsubsection}{\thesubsection .\arabic{subsubsection}}
   \renewcommand{\thefigure}{\thesection.\arabic{figure}}
   \renewcommand{\thetable}{\thesection.\arabic{table}}
}

\PassOptionsToPackage{hyperindex,breaklinks}{hyperref}
\usepackage{hyperref}
\usepackage{xcolor}
\definecolor{darkblue}{rgb}{0, 0, 0.5}
\hypersetup{colorlinks=true,citecolor=darkblue, linkcolor=darkblue, urlcolor=darkblue}

\usepackage[textsize=small]{todonotes}

\title{Identification of NMF by choosing maximum-volume basis vectors
}

\author{Qianqian Qi\\
   {\small\raggedright Hangzhou Dianzi University, China}\\
   \href{mailto:q.qi@hdu.edu.cn}{\texttt{q.qi@hdu.edu.cn}} 
\and Zhongming Chen\\
    {\small\raggedright Hangzhou Dianzi University, China}\\
\href{mailto:zmchen@hdu.edu.cn}{\texttt{zmchen@hdu.edu.cn}}
\and Peter G. M. van der Heijden\\
    {\small\raggedright Utrecht University, the Netherlands and University of Southampton, UK}\\
\href{mailto:p.g.m.vanderheijden@uu.nl}{\texttt{p.g.m.vanderheijden@uu.nl}}
    }

\predate{}
\postdate{}
\date{}
\date{\vspace{-5ex}}

\begin{document}
{\setstretch{.8}
\maketitle
\begin{abstract}

In nonnegative matrix factorization (NMF), minimum-volume-constrained NMF is a widely used framework for identifying the solution of NMF by making basis vectors as similar as possible. This typically induces sparsity in the coefficient matrix, with each row containing zero entries. Consequently, minimum-volume-constrained NMF may fail for highly mixed data, where such sparsity does not hold. Moreover, the estimated basis vectors in minimum-volume-constrained NMF may be difficult to interpret as they may be mixtures of the ground truth basis vectors. To address these limitations, in this paper we propose a new NMF framework, called maximum-volume-constrained NMF, which makes the basis vectors as distinct as possible. We further establish an identifiability theorem for maximum-volume-constrained NMF and provide an algorithm to estimate it. Experimental results demonstrate the effectiveness of the proposed method.
\end{abstract}
}



\section{Introduction}

Nonnegative matrix factorization (NMF) has been a workhorse method used in many areas, such as blind hyperspectral unmixing, blind sound separation, image processing, text mining, and signal processing \cite{lee1999learning, gillis2020nonnegative, SaberiMovahed2025Nonnegative}. It aims to approximate a nonnegative matrix as a product of two lower-rank nonnegative matrices: given a matrix $\bm{X}\in \Re^{I\times J}_{+}$ and a dimensionality $K \leq \text{min}\{I, J\}$, NMF searches for $\bm{M} \in \Re^{I\times K}_+$ and $\bm{H}\in \Re^{K\times J}_+$ such that $\bm{MH}$ approximates $\bm{X}$ as good as possible. Thus, NMF can be represented as
\begin{equation}\label{Eq: plain nmf}
\begin{split}
    ~~~  &  \bm{X} \approx \bm{M}\bm{H}\\
    \text{subject to} ~~~ & \bm{M}  \in \Re_{+}^{I\times K},~~~~\bm{H}  \in \Re_{+}^{K\times J},
\end{split}    
\end{equation}
where each column of $\bm{M}$ represents a basis vector, and each column of $\bm{H}$ contains the coefficients associated with the corresponding column (observation) of $\bm{X}$. In the noiseless setting, $\bm{X} = \bm{MH}$, which means $\bm{X}(:, j) \in \text{cone}(\bm{M})$. Given a matrix $\bm{A}$, $\text{cone}(\bm{A})$ is the convex cone generated by the columns of $\bm{A}$ \cite{gillis2020nonnegative}. That is, $\text{cone}(\bm{A}) = \{\sum_jy_j\bm{A}(:, j) | y_j \geq 0\}$, where $\bm{A}(:, j)$ is the $j$th column of $\bm{A}$.

\begin{figure}[H]
\centering
\begin{subfigure}[b]{0.3\linewidth}
\includegraphics[width=1\linewidth]{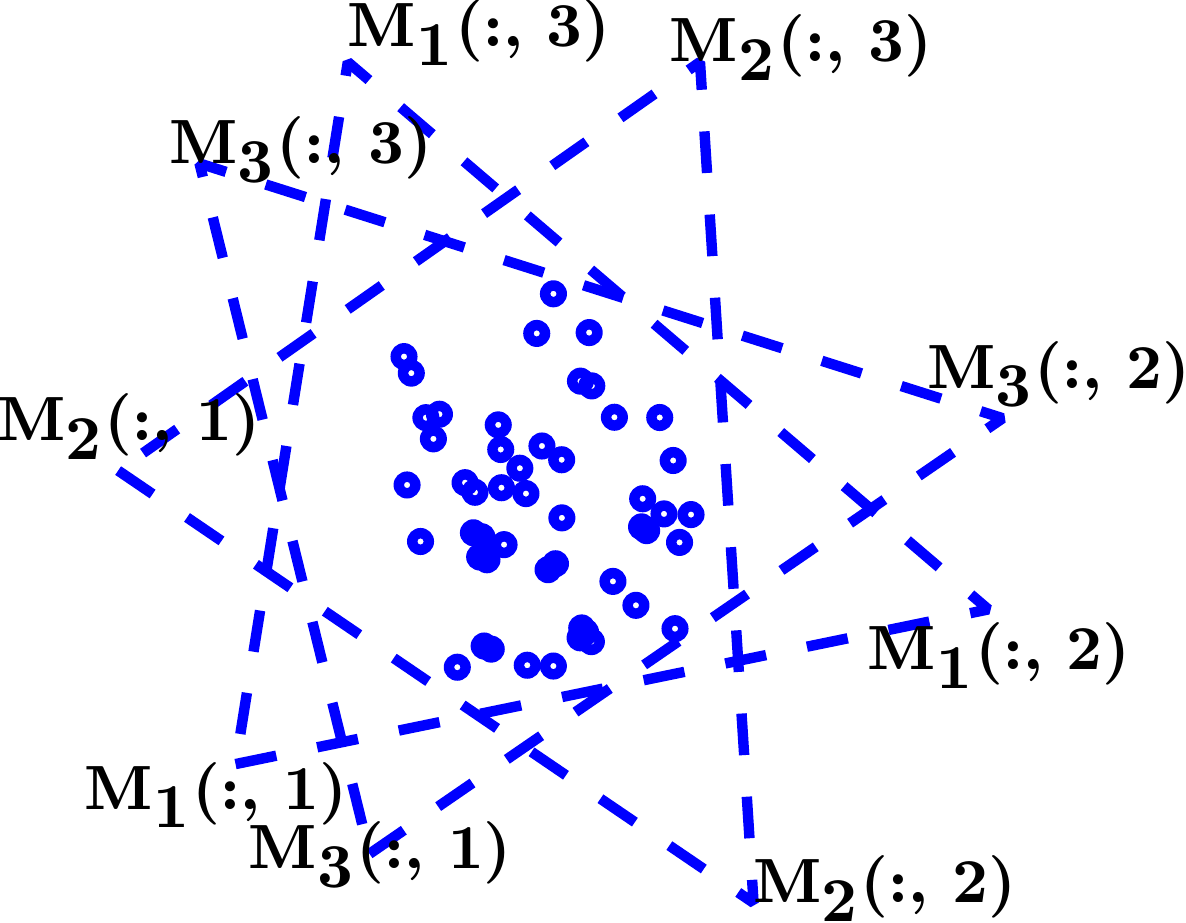}
 \caption{Nonuniqueness}\label{F: nonuniqueness}
 \end{subfigure}
\begin{subfigure}[b]{0.32\linewidth}
\includegraphics[width=1\linewidth]{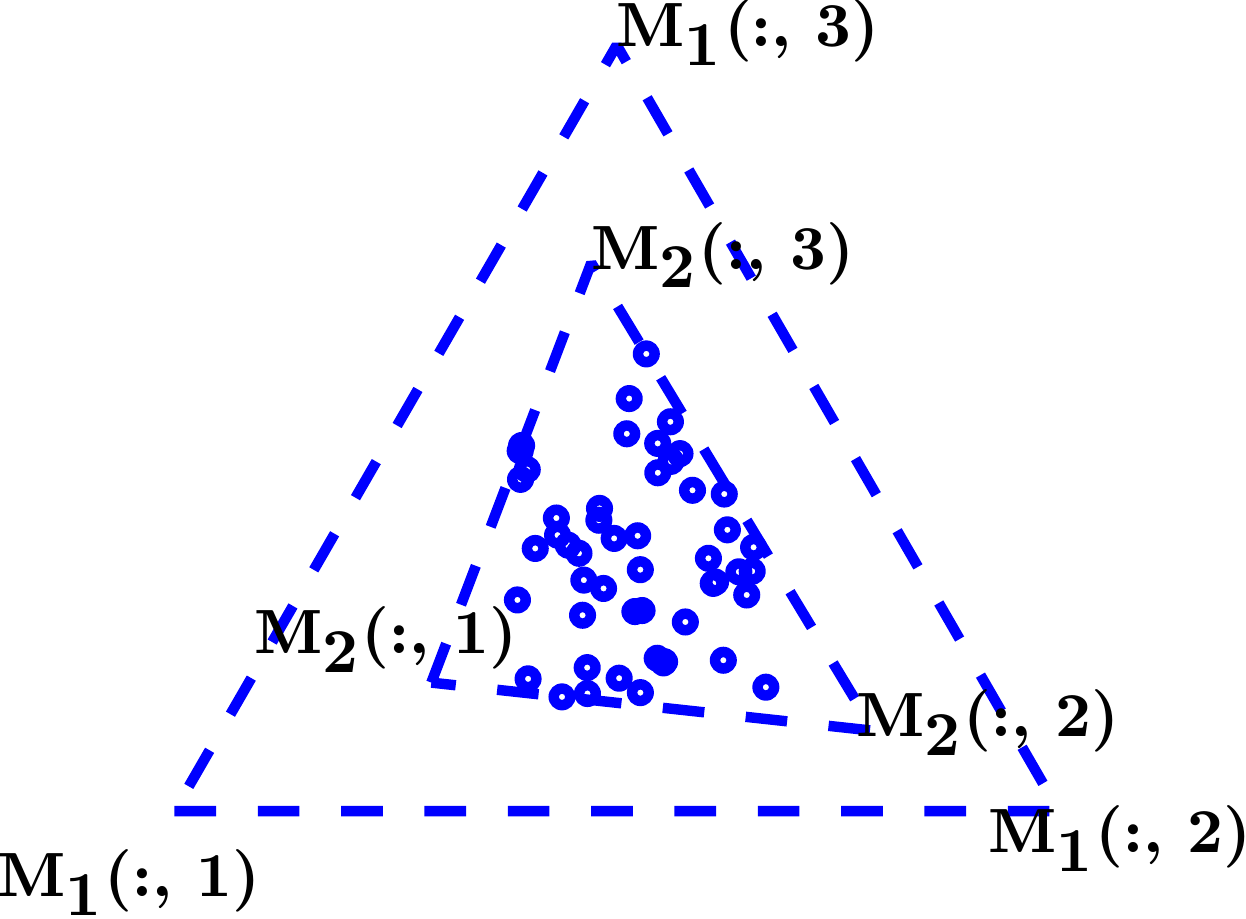}
 \caption{Minimum-volume NMF}\label{F: minvol}
 \end{subfigure}
\begin{subfigure}[b]{0.32\linewidth}
\includegraphics[width=1\linewidth]{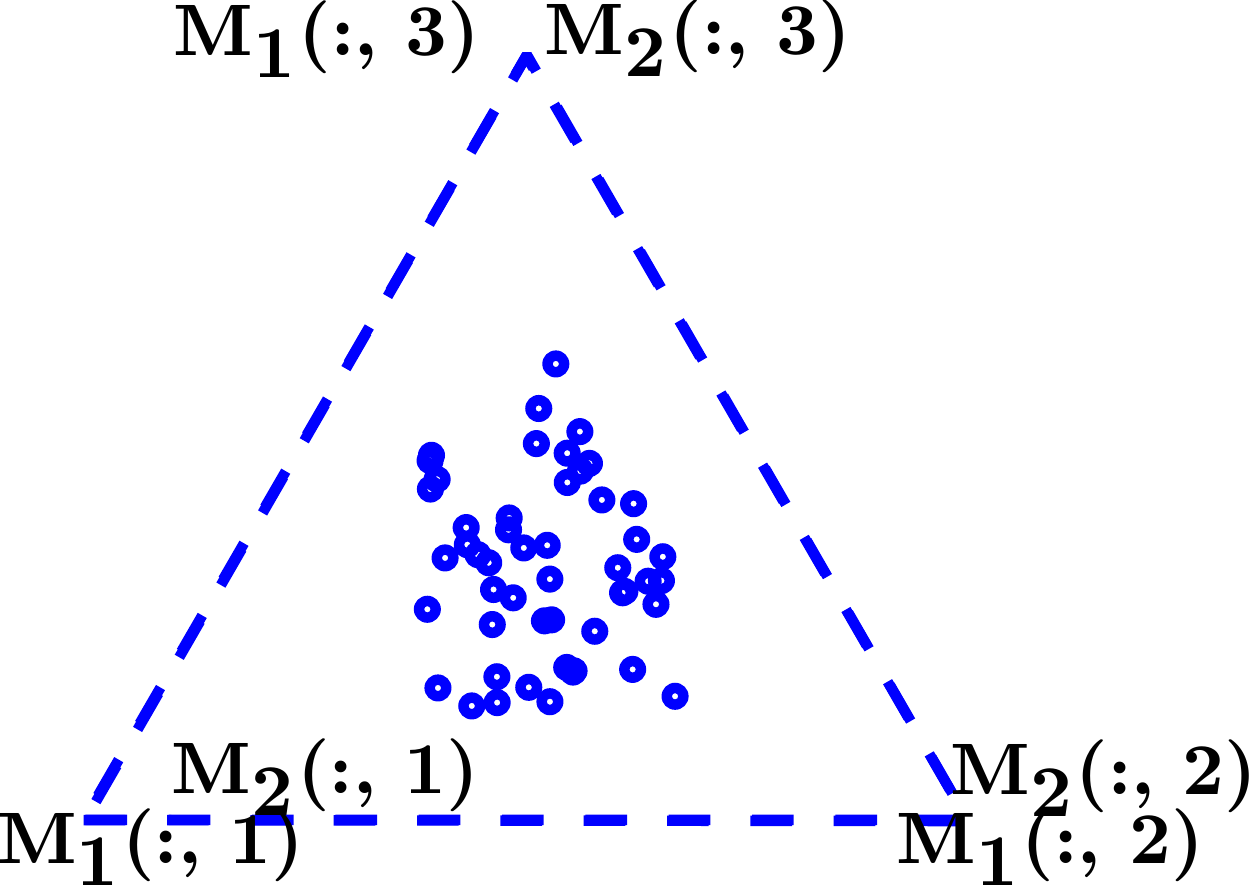}
 \caption{Maximum-volume NMF}\label{F: maxvol}
 \end{subfigure}
 \caption{A graphical view of columns of the matrix $\bm{X}$ (blue dots) and $\text{cone}(\bm{M})$ (dashed triangle), sliced by the unit-sum hyperplane $\bm{y}^T\bm{1} = 1$: (a) nonuniqueness; (b) minimum-volume-constrained NMF; (c) maximum-volume-constrained NMF.}
    \end{figure}

However, the solution to NMF (\ref{Eq: plain nmf}) is not unique \cite{gillis2020nonnegative, Fu2019Nonnegative, Abdolali2024Dual}.
Specifically, if $(\bm{M}, \bm{H})$ is a solution of (\ref{Eq: plain nmf}), $(\bm{M}\bm{S}, \bm{S}^{-1}\bm{H})$ is also a solution of (\ref{Eq: plain nmf}) for any invertible matrix $\bm{S} \in \Re^{K\times K}$, provided that both $\bm{\bm{M}\bm{S}}$ and $\bm{S}^{-1}\bm{H}$ are nonnegative. Figure~\ref{F: nonuniqueness} illustrates this non-uniqueness for $I = K = 3$, and $J = 50$ in the noiseless setting. Figure~\ref{F: nonuniqueness} is obtained by assuming that the viewer is located in the nonnegative orthant, facing the origin, and observing the unit-sum hyperplane $\bm{y}^T\bm{1} = 1$. In the figure, $\bm{M}_1$, $\bm{M}_2$, $\bm{M}_3$ represent three estimated basis matrices of NMF of $\bm{X}$.
Three dashed triangles represent \text{cone}($\bm{M}_1$), \text{cone}($\bm{M}_2$), and \text{cone}($\bm{M}_3$), respectively. As shown, blue dots (columns of $\bm{X}$) lie within all three triangles, indicating that $\bm{M}_1$, $\bm{M}_2$, and $\bm{M}_3$ are all feasible basis matrices of NMF for $\bm{X}$.

Minimum-volume-constrained NMF (MVC-NMF) is a widely used framework for identifying the basis matrix $\bm{M}$ and the coefficient matrix $\bm{H}$. In the noiseless setting, MVC-NMF can be formulated as \cite{fu2015}:
    \begin{equation}\label{Eq: minvolcriteria}
\begin{split}
    \text{min}\quad & \text{det}(\bm{M}^T\bm{M})\\
    \text{subject to} \quad  & \bm{X} = \bm{M}\bm{H}, \quad \bm{M}\in \Re^{I\times K}_+, \quad \bm{H}\in \Re^{K\times J}_+,
    \\& \bm{H}^T\bm{1} = \bm{1},
\end{split}    
\end{equation}
where $\text{det}(\bm{M}^T\bm{M})$ denotes the determinant of $\bm{M}^T\bm{M}$. The $\text{det}(\bm{M}^T\bm{M})$ is used because $\sqrt{\text{det}(\bm{M}^T\bm{M})}/K!$
corresponds to the volume of the simplex formed by the origin and the columns of $\bm{M}$ \cite{leplat2020, fu2016robust}. Given a matrix $\bm{A}$, the convex hull of its columns, denoted by $\text{conv}(\bm{A})$, is the sets of all convex combination of its columns, i.e., $\text{conv}(\bm{A}) = \{\sum_jy_j\bm{A}(:, j) | y_j \geq 0, \sum_jy_j = 1\}$; if the columns $\bm{A}(:, j)$, $j = 1, \cdots, J$ are affine independent, then $\text{conv}(\bm{A})$ is a simplex. The constraint $\bm{H}^T\bm{1} = \bm{1}$ in \eqref{Eq: minvolcriteria} can be replaced by $\bm{H}\bm{1} = \bm{1}$ or $\bm{M}^T\bm{1} = \bm{1}$ \cite{leplat2020, fu2018identifiability}.

Minimum-volume-constrained NMF is associated with sparsity in the coefficient matrix 
$\bm{H}$. An artificial dataset is generated by using simulated matrices $\bm{M}$ and $\bm{H}$, i.e., $\bm{X} = \bm{MH}$, where $I = 9$, $K = 3$, and $J = 50$. Figure~\ref{F: minvol} illustrates the sparsity of the coefficient matrix estimated by minimum-volume-constrained NMF. In the figure,
$\bm{M}_1$ denotes the ground truth basis matrix and $\bm{M}_2$ denotes the estimated basis matrix by minimum-volume-constrained NMF. For visualization convenience, $\bm{M}_1(:, 1)$, $\bm{M}_1(:, 2)$, and $\bm{M}_1(:, 3)$ are plotted as the vertices of an equilateral triangle; however, under the nonnegative orthant view, they do not necessarily form an equilateral triangle. Under the minimum-volume criterion (see the dashed triangle formed by $\bm{M}_2$), there are data points, i.e., columns of $\bm{X}$, on each boundary of the $\text{cone}(\bm{M}_2)$. As a result, every row of the estimated coefficient matrix $\bm{H}_2$ contains zero entries \cite{Zhou2012Minimum}.
Furthermore, \cite{fu2015, leplat2020, fu2018identifiability} showed that the solution to (\ref{Eq: minvolcriteria}) is unique when $\bm{H}$ is satisfied with the so-called sufficiently scattered condition (SSC), which requires every row of $\bm{H}$ to contain $K-1$ zero entries. To summarize, the success of minimum-volume-constrained NMF relies on sparsity in the coefficient matrix $\bm{H}$. 

While the minimum-volume assumption is appropriate in certain scenarios, it can be violated for highly mixed data, where the coefficient matrix $\bm{H}$ does not contain zero entries in every row. As illustrated in Figure~\ref{F: minvol}, the basis matrix $\bm{M}_2$ estimated by minimum-volume-constrained NMF cannot recover the ground-truth basis matrix $\bm{M}_1$. In addition, minimum-volume-constrained NMF minimizes $\det(\bm{M}^\top \bm{M})$, which makes the basis vectors as similar as possible. Consequently, the estimated basis vectors may be difficult to interpret. Moreover, the basis vectors estimated using minimum-volume-constrained NMF may remain mixtures of the ground truth basis vectors as shown in Figure~\ref{F: minvol}.

In this paper, we propose a new framework for NMF, coined maximum-volume-constrained NMF (MAV-NMF). In contrast to minimum-volume-constrained NMF, MAV-NMF maximizes the volume of $\bm{M}$. The basis vectors of this representation of NMF is more easily interpretable and unmixed and it is suitable for highly mixed data where the coefficient matrix does not contain zero entries in every row, as the estimated basis vectors are chosen to be as
distinct as possible. We further establish an identifiability theorem for MAV-NMF and develop a strategy for estimating it.
For the same artificial dataset as in Figure~\ref{F: minvol}, Figure~\ref{F: maxvol} demonstrates the effectiveness of MAV-NMF using the algorithm that will be introduced in Section~\ref{S: mavnmf}. As can be seen from the figure, the basis matrix $\bm{M}_2$ estimated by MAV-NMF coincides with the ground-truth basis matrix $\bm{M}_1$. We note that the maximum-volume-constrained NMF proposed in this paper is fundamentally different from \cite{Tatli2021Polytopic, thanh2026maximum}. Although the term maximum-volume in the context of NMF is used in \cite{Tatli2021Polytopic, thanh2026maximum}, but \cite{Tatli2021Polytopic, thanh2026maximum} maximize the volume of $\bm{H}$, which is similar to minimizing the volume of $\bm{M}$.

The paper is built up as follows. Section~\ref{S: Unique theorem} presents the MAV-NMF model and a condition guaranteeing its identifiability. In Section~\ref{S: mavnmf}, we develop a strategy for estimating MAV-NMF. Numerical experiments to illustrate
the performance of MAV-NMF are in Sections~\ref{S: Numerical experiments}. Finally, in Section~\ref{S: Conclusion}, we conclude.

\section{Uniqueness theorem for maximum-volume-constrained NMF}\label{S: Unique theorem}

In this section we propose maximum-volume-constrained nonnegative matrix factorization (MAV-NMF) and an uniqueness theorem for MAV-NMF. Most uniqueness theorems are established under the noiseless setting, which we assume in this section. Formally, the uniqueness or identifiability of the solution of nonnegative matrix factorization (NMF) is defined as follows.
\begin{definition}
\cite{gillis2020nonnegative, huang2014non} The NMF solution ($\bm{M}$, $\bm{H}$) of $\bm{X}$ is said to be essentially unique if and only if any other NMF solution ($\tilde{\bm{M}}$, $\tilde{\bm{H}}$) can be expressed as
    \begin{equation*}
        \tilde{\bm{M}} = \bm{M}(\bm{\Gamma}\bm{\Sigma}) \text{ and } \tilde{\bm{H}} = (\bm{\Gamma}\bm{\Sigma})^{-1}\bm{H},
    \end{equation*}
    where $\bm{\Gamma}$ is a permutation matrix and $\bm{\Sigma}$ is a diagonal matrix with positive diagonal entries.
\end{definition}
\noindent Here $\bm{\Gamma}$ allows for switching in the order of the columns of $\bm{M}$ and the rows of $\bm{H}$, and $\bm{\Sigma}$ allows for the adjustment of the vector sizes. 

In contrast to minimum-volume-constrained NMF (MVC-NMF) (\ref{Eq: minvolcriteria}), MAV-NMF maximizes the determinant of $\bm{M}^T\bm{M}$ instead of minimizing it. MAV-NMF can be formulated as:
    \begin{equation}\label{E: maxvolcriteria}
\begin{split}
    \text{max}\quad & \text{det}(\bm{M}^T\bm{M})\\
    \text{subject to} \quad  & \bm{X} = \bm{M}\bm{H},\quad \bm{M}\in \Re^{I\times K}_+, \quad \bm{H}\in \Re^{K\times J}_+,
    \\& \bm{H}^T\bm{1} = \bm{1}.
\end{split}    
\end{equation}
As in MVC-NMF, the constraint $\bm{H}^T\bm{1} = \bm{1}$ in \eqref{E: maxvolcriteria} is replaced by $\bm{H}\bm{1} = \bm{1}$ or $\bm{M}^T\bm{1} = \bm{1}$.

\begin{figure}[h]
\centering
 \begin{subfigure}[b]{0.23\linewidth}
\includegraphics[width=1\linewidth]{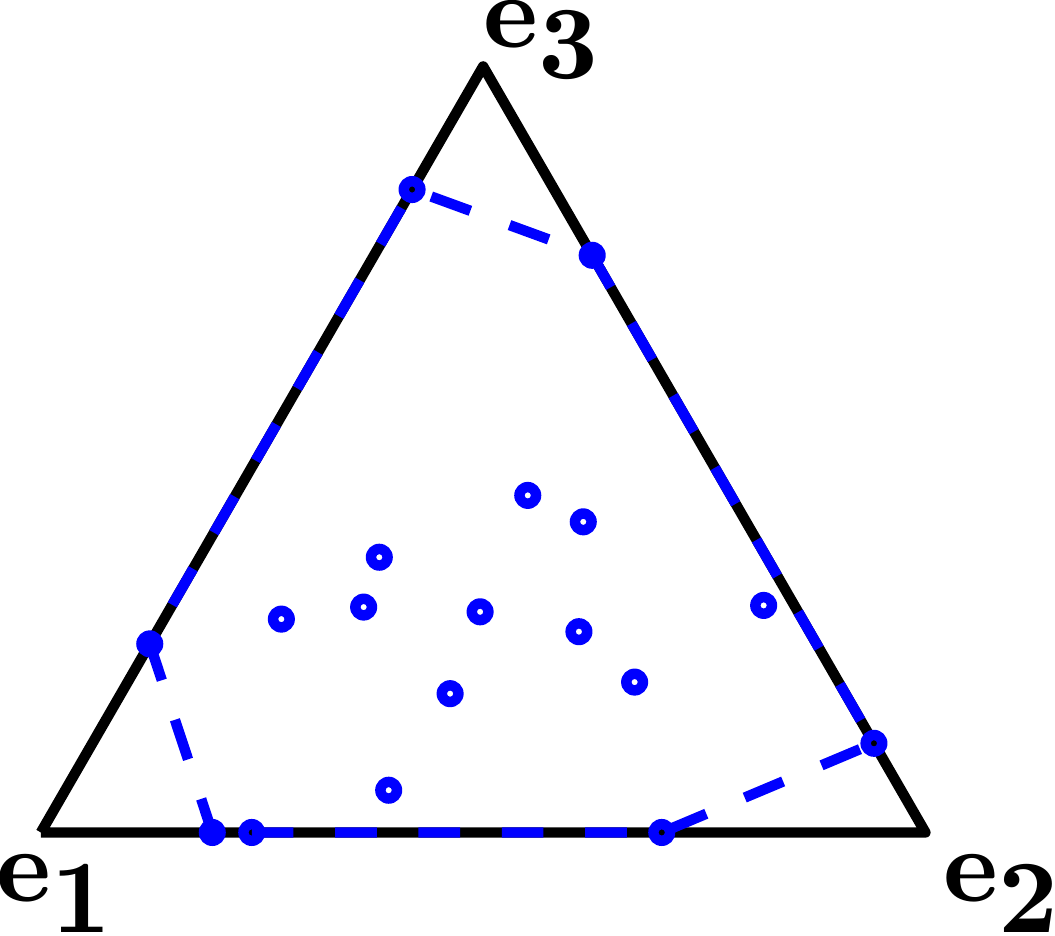}
\caption{MAV-NMF
}\label{F: maxvolm}
\end{subfigure}
  \begin{subfigure}[b]{0.23\linewidth}
\includegraphics[width=1\linewidth]{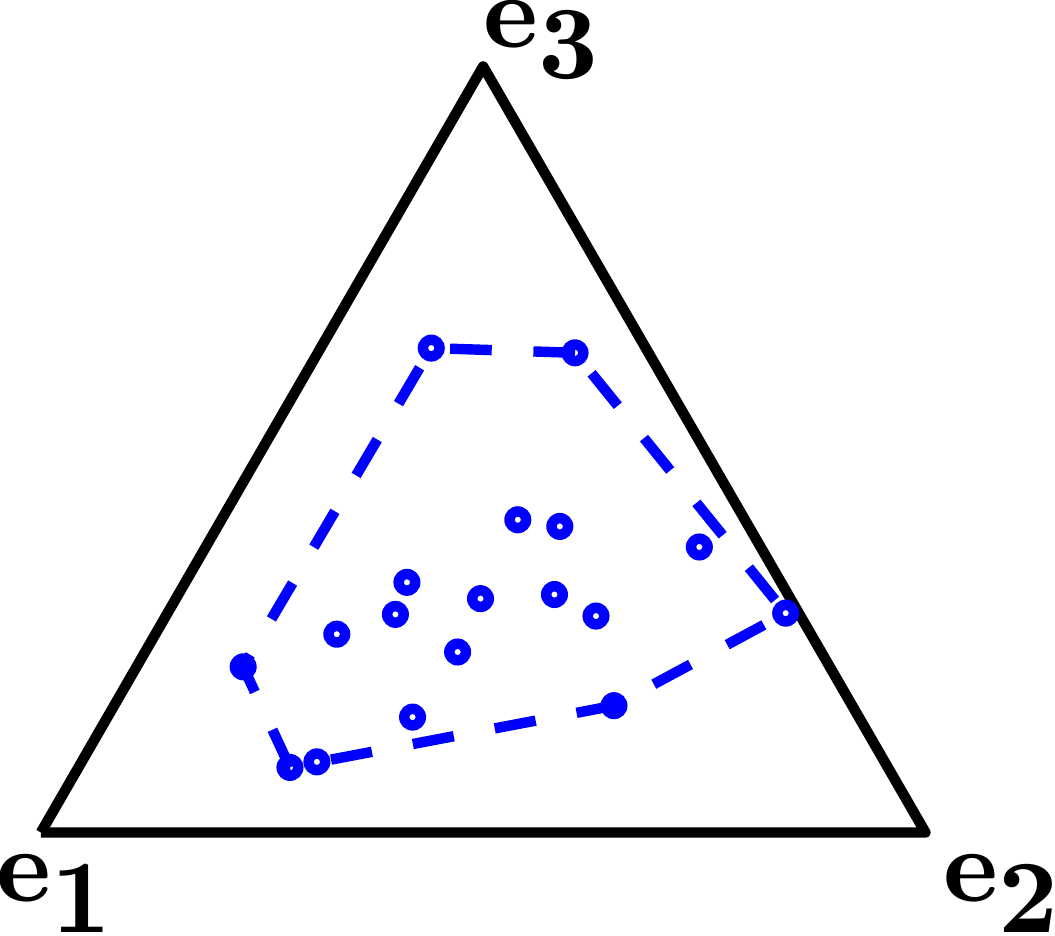}
\caption{MVC-NMF
}\label{F: minvolm}
\end{subfigure}
\begin{subfigure}[b]{0.23\linewidth}
\includegraphics[width=1\textwidth]{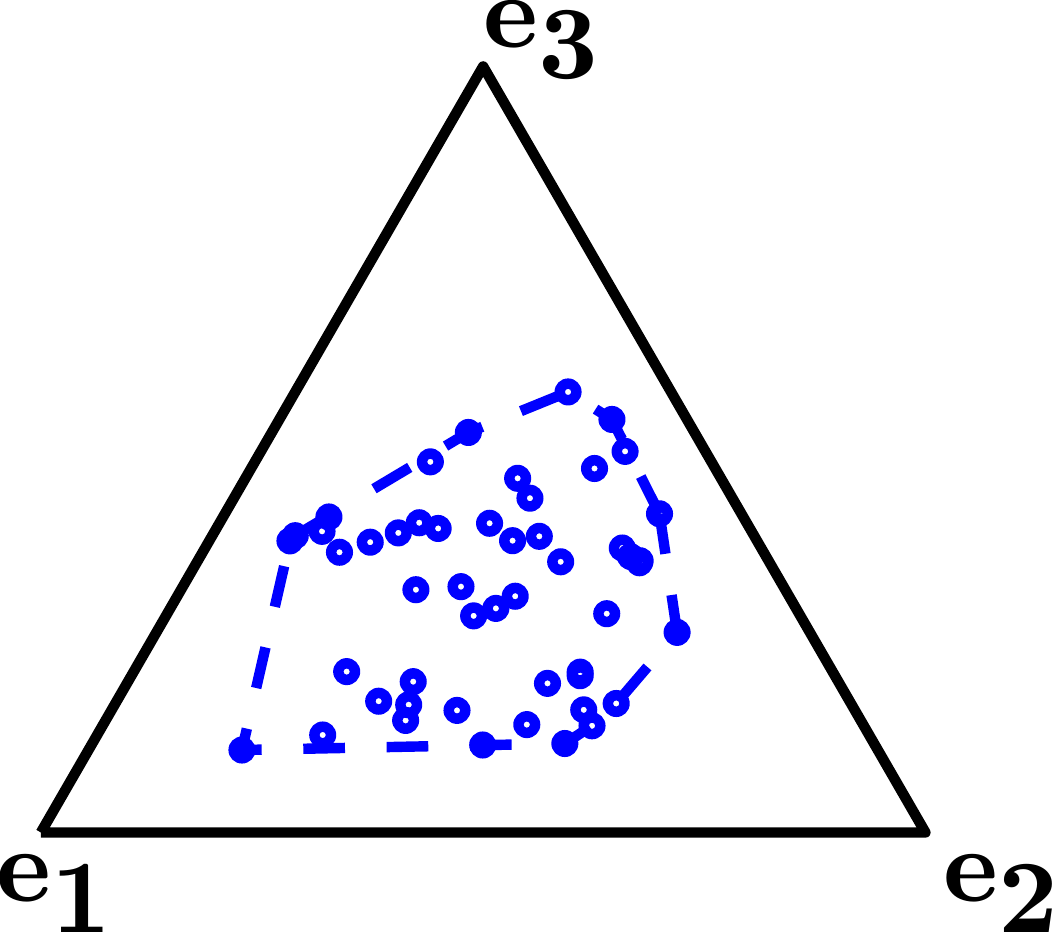}
 \caption{MAV-NMF
 }\label{F: maxvolh}
 \end{subfigure}
 \begin{subfigure}[b]{0.23\linewidth}
\includegraphics[width=1\textwidth]{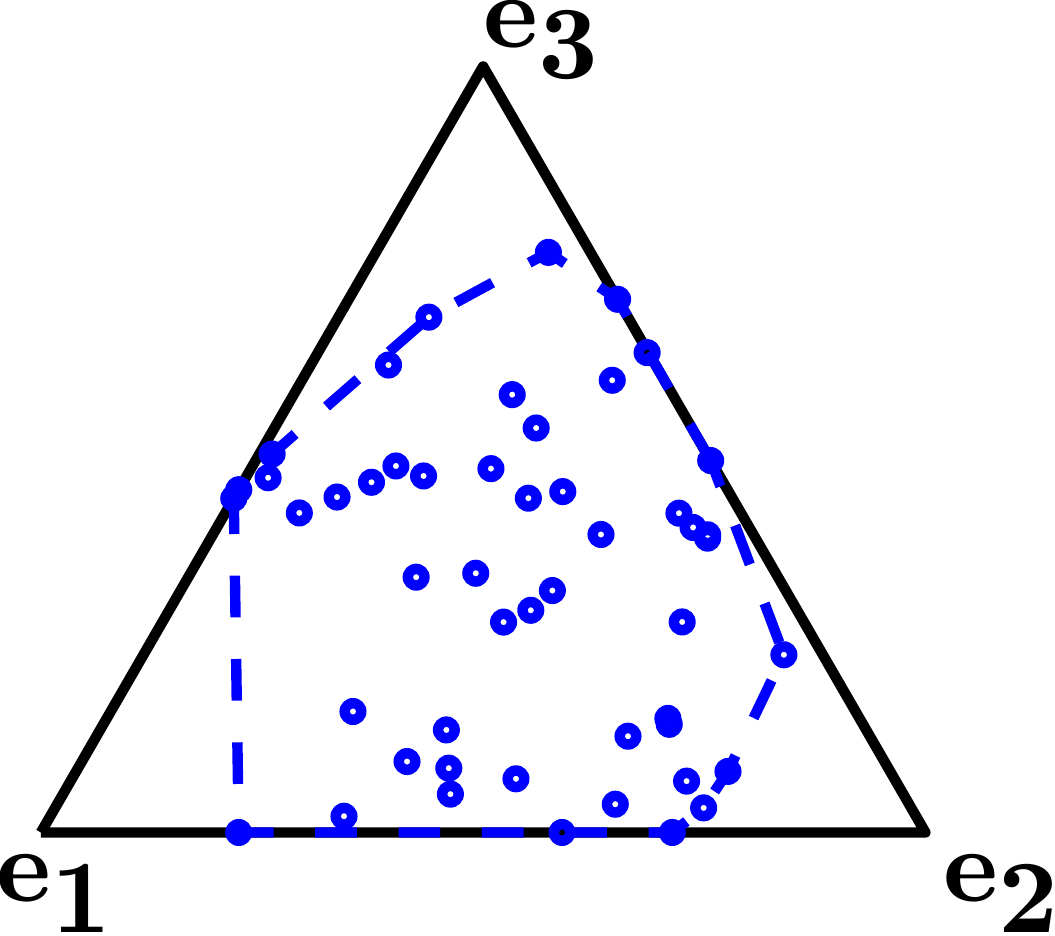}
 \caption{MVC-NMF
 }\label{F: minvolh}
 \end{subfigure}
 \caption{(a, b): A graphical representation of rows of the estimated basis matrix $\bm{MS}$ (blue dots), $\text{cone}(\bm{e}_1, \bm{e}_2, \bm{e}_3)$ (solid triangle) with $\bm{e}_1 = (1, 0, 0)^T$, $\bm{e}_2 = (0, 1, 0)^T$, and $\bm{e}_3 = (0, 0, 1)^T$, and $\text{cone}((\bm{MS})^T)$ (dashed polygon), sliced by the unit-sum hyperplane $\bm{y}^T\bm{1} = 1$; (c, d): A graphical representation of columns of the estimated coefficient matrix $\bm{S}^{-1}\bm{H}$ (blue dots), $\text{cone}(\bm{e}_1, \bm{e}_2, \bm{e}_3)$ (solid triangle), and $\text{cone}(\bm{S}^{-1}\bm{H})$ (dashed polygon), sliced by the unit-sum hyperplane $\bm{y}^T\bm{1} = 1$.}\label{F: maxminvolmh}
    \end{figure}

Maximizing $\text{det}(\bm{M}^T\bm{M})$ is equivalent to minimizing $\text{det}(\bm{H}\bm{H}^T)$. This can be seen by making use of a transformation matrix $\bm{S}$. Suppose that $(\bm{M}, \bm{H})$ is a solution to the NMF model (\ref{Eq: plain nmf}). Any solution of \eqref{Eq: plain nmf} can be written as $\tilde{\bm{M}} = \bm{M}\bm{S}$ and $\tilde{\bm{H}} = \bm{S}^{-1}\bm{H}$, where $\bm{S}$ is an invertible matrix. We have
\begin{equation}\label{Eq: maxmtominh}
    \text{max}~\det(\tilde{\bm{M}}^\top \tilde{\bm{M}}) \Longleftrightarrow \text{max}~ \det(\bm{S})^2  \Longleftrightarrow \text{min}~\det(\bm{S})^{-2}  \Longleftrightarrow \text{min}~\det(\tilde{\bm{H}}\tilde{\bm{H}}^\top).
\end{equation}
Figure~\ref{F: maxminvolmh} gives a simple illustration of this relationship for $K=3$. Figures~\ref{F: maxvolm} and~\ref{F: maxvolh} show 
$\text{cone}(\bm{(\bm{MS})^T})$ and 
$\text{cone}(\bm{S}^{-1}\bm{H})$ for MAV-NMF, respectively, while Figures~\ref{F: minvolm} and~\ref{F: minvolh} show 
$\text{cone}(\bm{(\bm{MS})^T})$ and 
$\text{cone}(\bm{S}^{-1}\bm{H})$ for MVC-NMF, respectively. As expected, 
$\text{cone}(\bm{(\bm{MS})^T})$ is larger for MAV-NMF (Figure~\ref{F: maxvolm}) than for MVC-NMF (Figure~\ref{F: minvolm}), whereas 
$\text{cone}(\bm{S}^{-1}\bm{H})$ is smaller for MAV-NMF (Figure~\ref{F: maxvolh}) than for MVC-NMF (Figure~\ref{F: minvolh}).

A sufficient condition of the uniqueness of the solution of MVC-NMF relies on the sufficiently scattered condition (SSC) in the coefficient matrix $\bm{H}$ \cite{fu2015, leplat2020, fu2018identifiability}. In contrast, for MAV-NMF, we impose the SSC on the transpose of the basis matrix, $\bm{M}^T$.
$\bm{M}^T$ is said to satisfy SSC if  \cite{gillis2020nonnegative, fu2015, leplat2020, fu2018identifiability, huang2014non}: 

(1) SSC1: $\mathbb{C} \subseteq \text{cone}(\bm{M}^T)$, where $\text{cone}(\bm{M}^T) = \{\sum_iy_i\bm{M}(i, :)^T | y_i \geq 0\}$, with $\bm{M}(i, :) \in \Re_{+}^{1\times K}$ denoting the $i$th row of $\bm{M}$; moreover, $\mathbb{C} = \{\bm{y} \in \Re_{+}^K | \bm{1}^T\bm{y} \geq \sqrt{K - 1}||\bm{y}||_2\}$, which is a second-order cone; 

(2) SSC2: $\text{cone}^*(\bm{M}^T) \cap  bd\mathbb{C}^* = \{\alpha\bm{e}_k | \alpha \geq 0, k = 1, \cdots, K\}$, where the dual cone of $\bm{M}^T$ is $\text{cone}^*(\bm{M}^T) = \{\bm{y} | \bm{M}\bm{y} \geq 0\}$, the dual cone of $\mathbb{C}$ is $\mathbb{C}^* = \{\bm{y} \in \Re^K | \bm{1}^T\bm{y} \geq ||\bm{y}||_2\}$, and $bd\mathbb{C}^*$ is the boundary of $\mathbb{C}^*$.

Now, we present an identifiability theorem for the solution of MVA-NMF, which relies on the SSC in the transpose of the basis matrix: $\bm{M}^T$.
\begin{theorem}\label{theorem: maxvol}
Assume that (1) $\text{rank}(\bm{M}) = \text{rank}(\bm{H}) = K$, (2) $\bm{M}^T$ satisfies SSC. Then MAV-NMF uniquely identifies $\tilde{\bm{M}}$ and $\tilde{\bm{H}}$ up to permutation. That is, any optimal solution $(\bm{M}_{\#}, \bm{H}_{\#})$ of MAV-NMF (\ref{E: maxvolcriteria}) can be expressed as:
\begin{equation*}
    \bm{M}_{\#} = \tilde{\bm{M}}\bm{\Gamma} \text{ and } \bm{H}_{\#} = \bm{\Gamma}^{-1}\tilde{\bm{H}}
\end{equation*}
where $\bm{\Gamma}$ is a permutation matrix.
\end{theorem}
\noindent The proof of Theorem~\ref{theorem: maxvol} follows arguments similar to those developed for MVC-NMF in \cite{fu2015, leplat2020, fu2018identifiability} and is provided in the supplementary materials. 

In MAV-NMF, the SSC is imposed on the transpose $\bm{M}^T$ of the basis matrix rather than on the coefficient matrix $\bm{H}$, as in MVC-NMF. The SSC is closely related to sparsity: if $\bm{M}^T$ satisfies SSC, each column of $\bm{M}$ contains at least $K-1$ zero entries \cite{gillis2020nonnegative, huang2014non}. Consequently, MAV-NMF promotes sparsity in the basis matrix $\bm{M}$ instead of in the coefficient matrix $\bm{H}$, which can be advantageous for highly mixed data and leads to basis vectors that are more easily interpretable than those in MVC-NMF. In Figure~\ref{F: maxvolm}, $(\bm{MS})^T$ satisfies SSC, and thus the solution of MAV-NMF is unique.

\section{A simple approach for solving maximum-volume-constrained NMF}\label{S: mavnmf}

In general, $\bm{X} - \bm{M}\bm{H} \neq \bm{0}$. The objective function of maximum-volume-constrained NMF (MAV-NMF) can be formulated as
\begin{equation}\label{Eq: nmfobjneg}
\begin{split}
     \text{min } ~~~ &||\bm{X} - \bm{M}\bm{H}||_F^2 - \lambda \text{logdet}(\bm{M}^T\bm{M} + \delta \bm{I})\\
 \text{subject to } ~~~   &  \bm{M}  \in \Re_{+}^{I\times K},~~~~\bm{H}  \in \Re_{+}^{K\times J}, ~~~~\bm{H}^T\bm{1} = \bm{1},
\end{split}    
\end{equation}
where $||\bm{X} - \bm{M}\bm{H}||_F^2$ is the data-fitting term and $\log\det(\bm{M}^T\bm{M} + \delta \bm{I})$ is the volume regularization term, with $\delta > 0$ and $\lambda > 0$ \cite{fu2016robust, Leplat2019rank}. For a given matrix $\bm{A} \in \mathbb{R}^{I\times J}$, the Frobenius norm $||\bm{A}||_F^2$ is defined as $||\bm{A}||_F^2 = \sqrt{\sum_{i}\sum_{j} \bm{A}(i,j)^2}$. The parameter $\lambda$ controls the trade-off between the data-fitting term and the volume regularizer. The main difference between MAV-NMF and the minimum-volume-constrained NMF (MVC-NMF) in \cite{Leplat2019rank} lies in the sign before the volume regularization term: a negative sign in \eqref{Eq: nmfobjneg} favors larger values of $\log\det(\bm{M}^T\bm{M} + \delta \bm{I})$. The constraint $\bm{H}^T\bm{1} = \bm{1}$ in \eqref{Eq: nmfobjneg} can be replaced by $\bm{H}\bm{1} = \bm{1}$ or $\bm{M}^T\bm{1} = \bm{1}$. Note that maximizing $\text{logdet}(\bm{M}^T\bm{M})$ avoids $\bm{M}^T\bm{M}$ to be singular. Here we use $\text{logdet}(\bm{M}^T\bm{M} + \delta \bm{I})$ instead of $\text{logdet}(\bm{M}^T\bm{M})$ because we will replace the negative volume term with a positive term, see (\ref{Eq: nmfobjnegtransfer}). 

The objective function in~\eqref{Eq: nmfobjneg} is non-convex. As in \cite{gillis2020nonnegative, Tatli2021Polytopic, Leplat2019rank, ang2019algorithms, ang2025sum}, we adopt an alternating optimization scheme over $\bm{M}$ and $\bm{H}$: iteratively updating $\bm{M}$ for fixed $\bm{H}$ and updating $\bm{H}$ for fixed $\bm{M}$. The main difficulty in solving (\ref{Eq: nmfobjneg}) with respect to $\bm{M}$ lies in the term $- \lambda \text{logdet}(\bm{M}^T\bm{M} + \delta \bm{I})$ \cite{Tatli2021Polytopic, thanh2026maximum}. Since the $\text{logdet}(\bm{Q})$ is concave, the first-order Taylor expansion is a majorizer of $\text{logdet}(\bm{Q})$ on the set of positive definite matrices $\bm{Q}$ and it is then possible to update $\bm{M}$ by minimizing the obtained
majorizer as in \cite{fu2016robust, Leplat2019rank}. However, since $-\text{logdet}(\bm{Q})$ is convex, it prevents us from using this strategy.

As discussed in the previous section (see \eqref{Eq: maxmtominh}), in the noiseless setting, maximizing $\det(\bm{M}^\top \bm{M})$ is equivalent to minimizing $\det(\bm{H}\bm{H}^\top)$. Motivated by this equivalence, we reformulate \eqref{Eq: nmfobjneg} as the following optimization problem:
\begin{equation}\label{Eq: nmfobjnegtransfer}
\begin{split}
     \text{min } ~~~ &||\bm{X} - \bm{M}\bm{H}||_F^2 + \lambda \text{logdet}(\bm{H}\bm{H}^T + \delta \bm{I})\\
 \text{subject to } ~~~   &  \bm{M}  \in \Re_{+}^{I\times K},~~~~\bm{H}  \in \Re_{+}^{K\times J}, ~~~~\bm{H}^T\bm{1} = \bm{1}.
\end{split}    
\end{equation}
In other words, by replacing $- \lambda \text{logdet}(\bm{M}^T\bm{M} + \delta \bm{I})$ in (\ref{Eq: nmfobjneg}) with $\lambda \text{logdet}(\bm{H}\bm{H}^T + \delta \bm{I})$  in (\ref{Eq: nmfobjnegtransfer}), the negative log-determinant term of $\bm{M}^T\bm{M} + \delta \bm{I}$ is effectively removed from the optimization problem. As in \eqref{Eq: nmfobjneg}, the constraint $\bm{H}^T\bm{1} = \bm{1}$ in \eqref{Eq: nmfobjnegtransfer} can be replaced by $\bm{H}\bm{1} = \bm{1}$ or $\bm{M}^T\bm{1} = \bm{1}$.

Problem (\ref{Eq: nmfobjnegtransfer}) can be solved by considering 
\begin{equation}\label{eq: objmin}
    \begin{split}
   \text{min}_{\bm{M}, \bm{H}}~&||\bm{X} - \bm{M}\bm{H}||_F^2 + \lambda \text{logdet}(\bm{M}^T\bm{M} + \delta \bm{I})\\
 \text{subject to } ~ & \bm{M}  \in \Re_{+}^{I\times K}, ~~~ \bm{H}  \in \Re_{+}^{K\times J}, ~~~\bm{M}\bm{1} = \bm{1},
    \end{split}
\end{equation}
making use of $\bm{X}^T = \bm{H}^T\bm{M}^T$. Similar to how the constraint 
$\bm{M}\bm{1} = \bm{1}$ in \eqref{eq: objmin} replaces $\bm{H}^T\bm{1} = \bm{1}$ in \eqref{Eq: nmfobjnegtransfer}, the alternative constraints 
$\bm{H}\bm{1} = \bm{1}$ and $\bm{M}^T\bm{1} = \bm{1}$ in \eqref{Eq: nmfobjnegtransfer} correspond to $\bm{M}^T\bm{1} = \bm{1}$ and $\bm{H}\bm{1} = \bm{1}$ in \eqref{eq: objmin}, respectively. In this paper, we use the algorithm in \cite{Leplat2019rank} which solved $ \text{min } ||\bm{X} - \bm{M}\bm{H}||_F^2 + \lambda \text{logdet}(\bm{M}^T\bm{M} + \delta \bm{I})$. We call the algorithm APFGM-LOGDET, short for alternative projected fast gradient methods for logdet regularized objective function. The algorithm from \cite{Leplat2019rank} is provided in Algorithm~1 and briefly outlined here. Given $\bm{M}$, the objective for $\bm{H}$ is
\begin{equation}\label{eq: objminh}
\min_{\bm{H} \geq 0}~||\bm{X} - \bm{MH}||_F^2.
\end{equation}
One updates $\bm{H}$ using a projected fast gradient method (PFGM) \cite{gillis2020nonnegative, Leplat2019rank}, which can incorporate constraints such as nonnegativity, column-sum-to-1, or row-sum-to-1. Given $\bm{H}$, the objective for $\bm{M}$ is $\text{min}_{\bm{M}\geq 0, \bm{M1} = \bm{1}}~||\bm{X} - \bm{MH}||_F^2+\lambda\text{logdet}(\bm{M}^T\bm{M}+\delta \bm{I})$. One updates $\bm{M}$ by applying PFGM to its strongly convex upper approximation:
\begin{equation}\label{eq: objminm}
    \text{min}_{\bm{M} \in \Re_+^{I\times K}, \bm{M1} = \bm{1}}~2\sum_{i = 1}^I\left(\frac{1}{2}\bm{M}(i, :)\bm{A}\bm{M}(i, :)^T -\bm{C}(i, :)\bm{M}(i, :)^T\right),
\end{equation}
where $\bm{A} = \bm{H}\bm{H}^T + \lambda(\bm{M}_0^T\bm{M}_0+\delta I)^{-1}$ and $\bm{C} = \bm{X}\bm{H}^T$. 

\begin{algorithm}\label{Alg: apfgmlogdet}

\caption{APFGM-LOGDET (\ref{eq: objmin}).} 

\KwIn{$\bm{X}$, $K$, $\lambda'$, and \it maxiter.}
\KwOut{$\bm{M} \ge 0$ and $\bm{H} \ge 0$.}

Generate initial matrices $\bm{M}^{(0)} \ge 0$ and $\bm{H}^{(0)} \ge 0$. Let $\lambda = \lambda'\frac{||\bm{X} - \bm{M}^{(0)}\bm{H}^{(0)}||_F^2}{|\text{logdet}(\bm{M}^T\bm{M} + \delta \bm{I})|}$.

\For{$t = 1, 2, \ldots, $\text{maxiter}}{
  PFGM is performed on (\ref{eq: objminh}) to update $\bm{H}$
  
  PFGM is performed on (\ref{eq: objminm}) to update $\bm{M}$
}
\end{algorithm}

\section{Numerical experiments}\label{S: Numerical experiments}

In this section, we evaluate the performance of maximum-volume-constrained NMF (MAV-NMF) on two artificial grain-size distribution datasets from \cite{ZHANG2020106656} in sedimentary geology, three artificial datasets simulated by us, a real-world human face images dataset from image processing, and a real-world time-allocation dataset from social science, where the three artificial datasets simulated by us are provided in the supplementary materials for saving space. For the artificial datasets, the ground truth basis and coefficient matrices are known, whereas for the real-world datasets, the ground truth is unknown. We compare MAV-NMF with minimum-volume-constrained NMF (MVC-NMF). All methods are implemented in MATLAB R2024b, and the code is available at \url{https://github.com/qianqianqi28/MAV-NMF}.

\subsection{Sedimentary geology: Grain-size distributions dataset}

\begin{figure}
\centering
 \begin{subfigure}[b]{0.32\linewidth}
\includegraphics[width=1\textwidth]{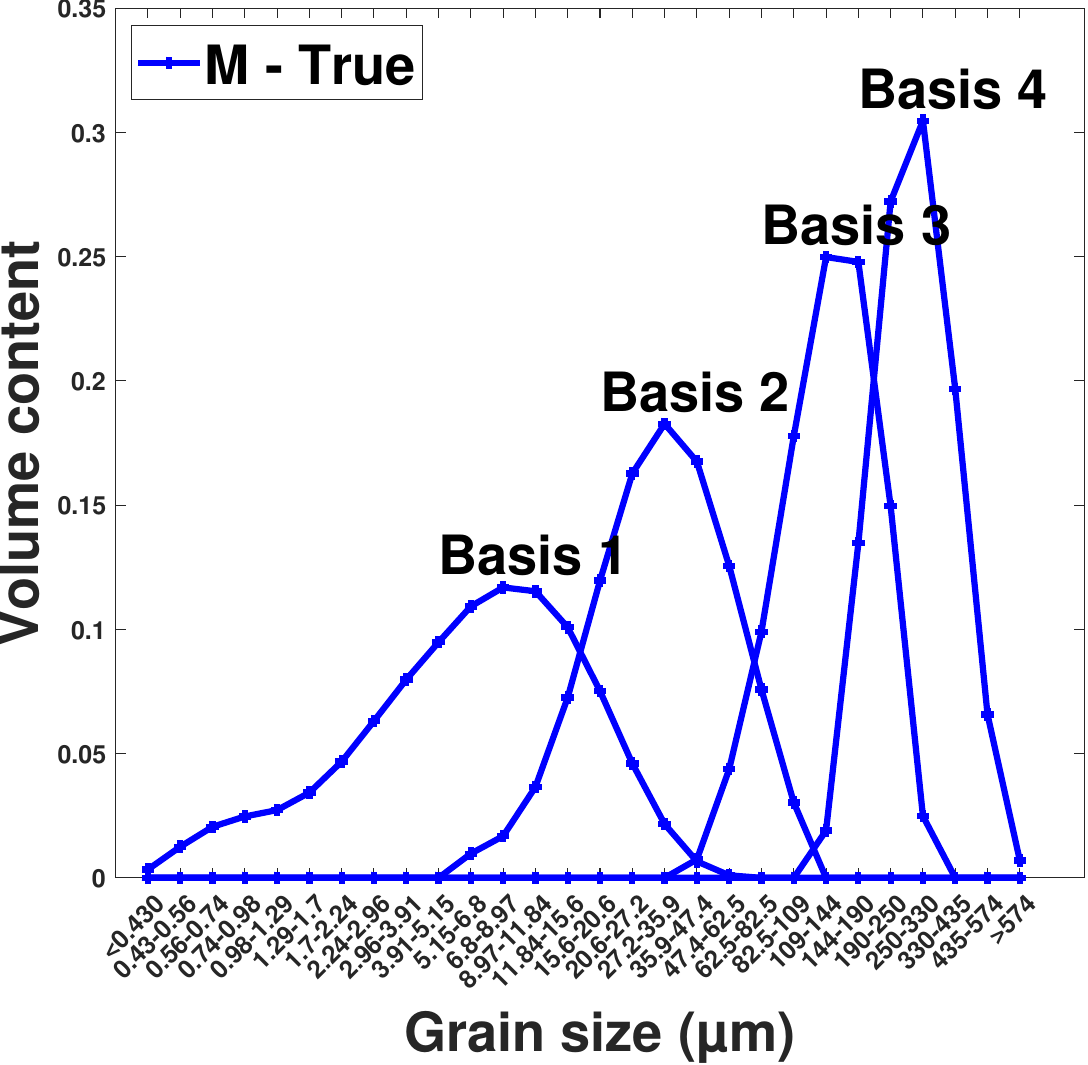}
 \caption{True basis vectors}\label{F: sgdtruebasis}
 \end{subfigure}
 \begin{subfigure}[b]{0.32\linewidth}
\includegraphics[width=1\textwidth]{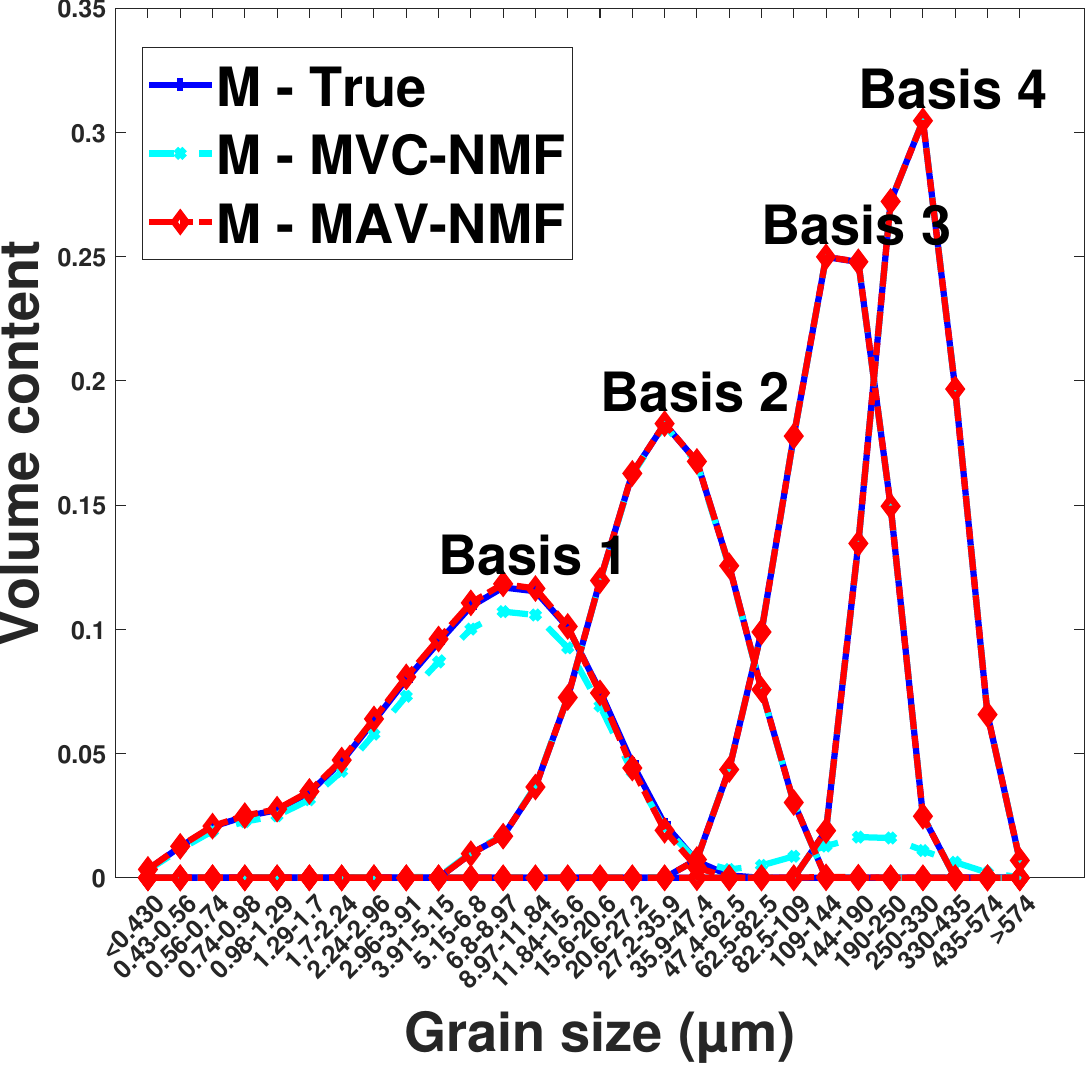}
 \caption{Test Dataset 1}\label{F: sgdminplainmaxmtest1}
 \end{subfigure}
 \begin{subfigure}[b]{0.32\linewidth}
\includegraphics[width=1\textwidth]{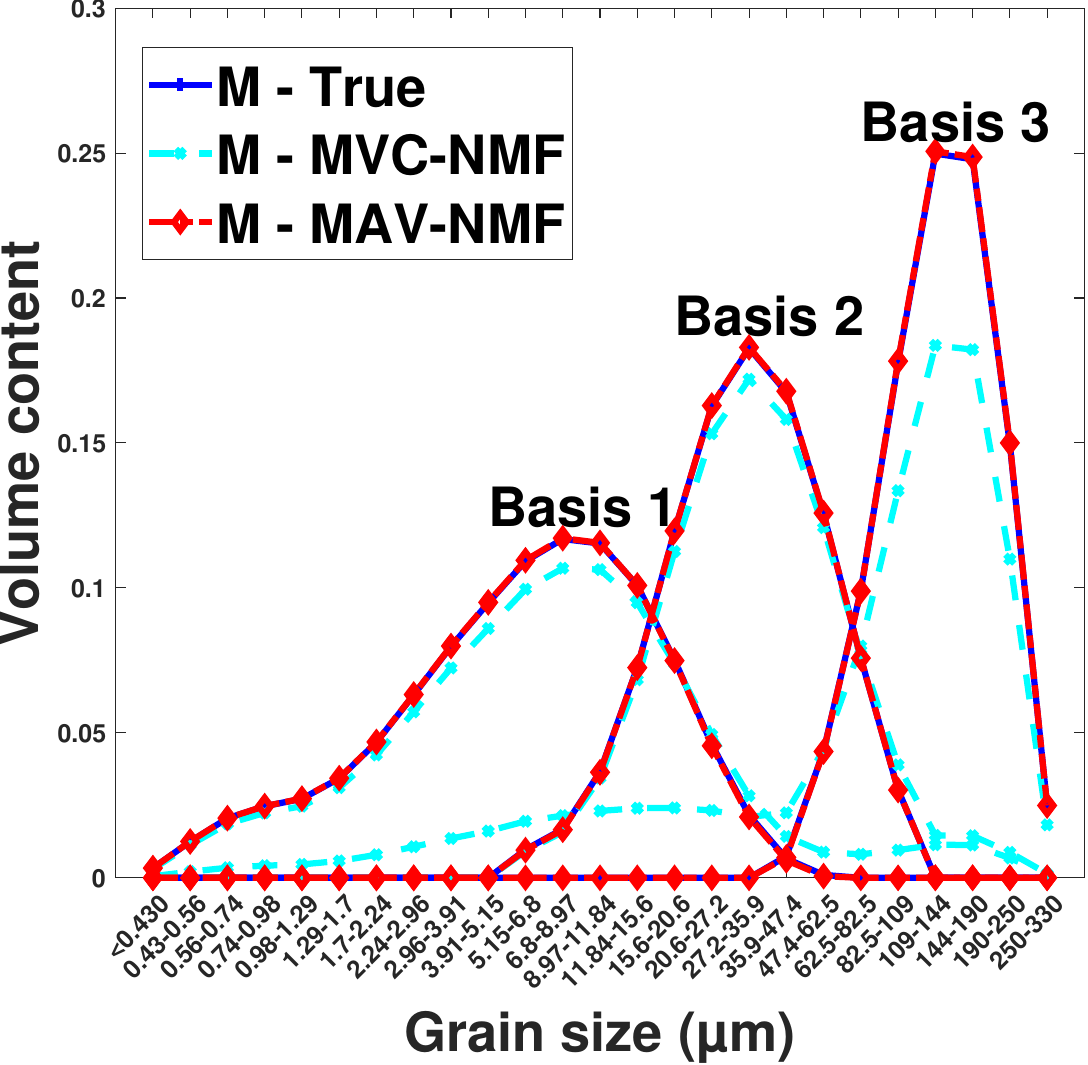}
 \caption{Test Dataset 2}\label{F: sgdminplainmaxmtest2}
 \end{subfigure}
 \caption{Artificial grain-size distributions
(GSDs) datasets from \cite{ZHANG2020106656} (a) The four ground truth basis vectors used from \cite{ZHANG2020106656}, which were from the calculated EMs from the grain-size data of the surface sediment in the South Yellow Sea by \cite{zhang2016end}; (b) Test Dataset 1; (c) Test Dataset 2.}\label{F: artificialdatasgd}
    \end{figure}

NMF can be used to unmix grain-size distribution (GSD) datasets, providing insights into sediment provenance, transport, and sedimentation processes \cite{ZHANG2020106656, paterson2015new, zhang2016end, van2018genetically, Liu2023Universal, moskalewicz2024identification, Lin2025Using, RENNY2026113384}. We use two artificial datasets from \cite{ZHANG2020106656}. Figure~\ref{F: sgdtruebasis} shows the four basis vectors, each with 28 values, used in \cite{ZHANG2020106656} to generate the two artificial datasets. These basis vectors were derived from grain-size data of surface sediments in the South Yellow Sea \cite{zhang2016end}.

For Test Dataset 1 with $I = 28$ and $J = 200$, the coefficient matrix of size $4 \times 200$ has a minimum value of 0.0003 and a maximum value of 0.8006; therefore, it contains no zero entries. Test Dataset 2 with $I = 25$ and $J = 200$ represents more highly mixed data generated from bases 1-3, with coefficients ranging from 0.0506 to 0.7939. The last three entries of bases 1-3 are zero; hence, they have been omitted in Test Dataset 2.

We apply MVC-NMF and MAV-NMF to Test Datasets 1 and 2. As shown in Figure~\ref{F: sgdminplainmaxmtest1}, MAV-NMF recovers basis vectors that are more consistent with the ground truth bases than those obtained by MVC-NMF. In particular, for MVC-NMF, the estimated basis~1 exhibits two peaks: one coincides with the peak of the ground truth basis~1, while the other lies between those of bases~3 and~4. A closer inspection of the estimated basis vectors shows that
\begin{equation*}
\begin{split}
     \text{estimated basis 1 by MVC-NMF} \approx & 0.9173 \times \text{ground truth basis 1} \\& 
     + 0.0490 \times \text{ground truth basis 3} \\&
     + 0.0320 \times \text{ground truth basis 4}.
\end{split}
\end{equation*}
Therefore, the estimated basis~1 produced by MVC-NMF does not correspond to a single ground truth basis but remains a mixture of multiple ground truth basis vectors. Moreover, the coefficient matrix estimated by MVC-NMF contains zero entries, which contradicts the true setting in which no coefficients are zero. These effects are even more pronounced for Test Dataset~2, which is more highly mixed than Test Dataset~1. 

Table~\ref{T: volsgd} reports $\log\det(\bm{M}^\top\bm{M} + \delta \bm{I})$, from which we observe that the volume obtained by MAV-NMF is larger than that obtained by MVC-NMF.

The three artificial
datasets simulated by us in the supplementary materials has similar results, namely MAV-NMF produces better results than MVC-NMF for highly mixed data.

\begin{table}
\caption{Artificial grain size distributions datasets; $\text{logdet}(\bm{M}^T\bm{M} + \delta \bm{I})$}
    \label{T: volsgd}
    \centering
    \begin{tabular}{c|c|c}
    \hline 
    Methods & Test Dataset 1 & Test Dataset 2\\
    \hline
 MVC-NMF &  -5.7100 & -4.9230  \\
  MAV-NMF & -5.6270 & -4.4340 \\
    \hline
    \end{tabular}
\end{table} 

\subsection{Image processing: Human face images dataset}

\begin{figure}[H]
\centering
  \begin{subfigure}[b]{0.45\linewidth}
\includegraphics[width=1\textwidth]{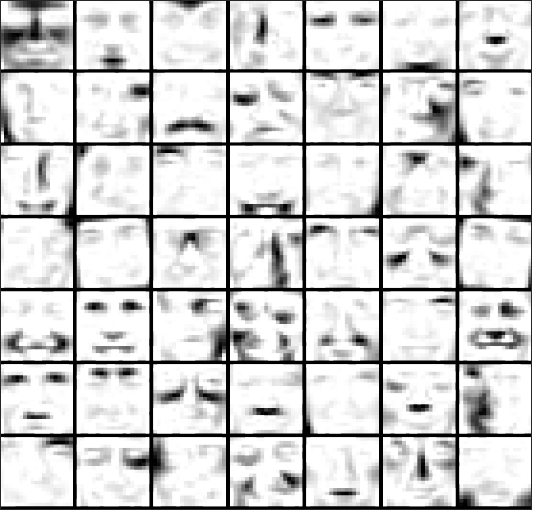}
 \caption{MVC-NMF}\label{F: cbclmin}
 \end{subfigure}
    \hspace{0.04\linewidth}
   \begin{subfigure}[b]{0.45\linewidth}
\includegraphics[width=\columnwidth]{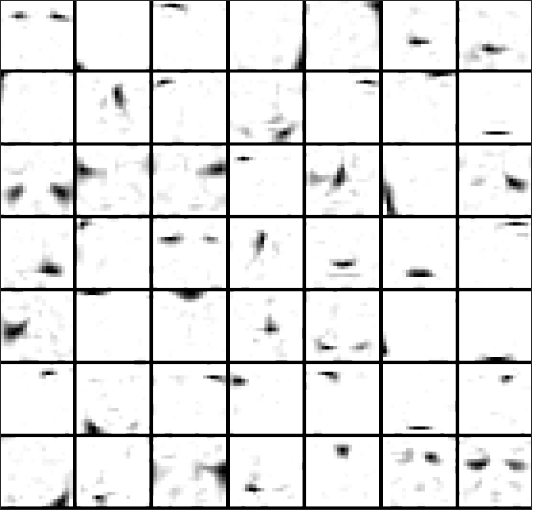}
    \caption{MAV-NMF}
    \label{F: cbclmax}
 \end{subfigure}
 \caption{CBCL face data set: (a) MVC-NMF; (b) MAV-NMF.}\label{F: cbcl}
    \end{figure}

We apply MVC-NMF and MAV-NMF to the CBCL face dataset of $J = 2,429$ face images, each consisting of $I = 19 \times 19$ pixels, i.e., $\bm{X} \in \mathbb{R}_+^{361 \times 2429}$ \cite{lee1999learning, gillis2020nonnegative}. For this dataset, in (\ref{eq: objmin}) we use the constraint $\bm{H}\bm{1} = \bm{1}$ instead of the constraint $\bm{M}\bm{1} = \bm{1}$. Following \cite{lee1999learning, gillis2020nonnegative}, we set the dimensionality to $K = 49$. Figure~\ref{F: cbclmin} and Figure~\ref{F: cbclmax} present the basis matrices $\bm{M}$ estimated by MVC-NMF and MAV-NMF, respectively. Each estimated basis matrix $\bm{M} \in \mathbb{R}_+^{361 \times 49}$ is reshaped into 49 individual images and displayed in a $7 \times 7$ grid, where positive values are displayed as black pixels. 

We observe that the basis obtained by MAV-NMF is substantially sparser than those produced by MVC-NMF, which is consistent with the theoretical analysis in Section~\ref{S: Unique theorem}. Moreover, the 49 basis images estimated by MAV-NMF contain multiple distinct versions of eyes, noses, and other facial parts, aligning better with intuitive facial parts. Table~\ref{T: volcbcl} reports $\log\det(\bm{M}^\top\bm{M} + \delta \bm{I})$, showing that MAV-NMF yields a larger volume than MVC-NMF.

\begin{table}[H]
\caption{Real-world CBCL dataset; $\text{logdet}(\bm{M}^T\bm{M} + \delta \bm{I})$}
    \label{T: volcbcl}
    \centering
    \begin{tabular}{c|c}
    \hline 
    Methods &CBCL \\
    \hline
 MVC-NMF &    -107.7660
  \\
  MAV-NMF &   -90.4210 \\
    \hline
    \end{tabular}
\end{table}

\subsection{Social science: Time allocation dataset}

We analyze a time-allocation dataset, where $I = 18$ corresponds to 18 activities (e.g., paid work, domestic work, and caring for household members) and $J = 30$ represents groups cross-classified by gender (male and female), age (12–24, 25–34, 35–49, 50–64, and $>65$), and survey year (1975, 1980, and 1985) \cite{mooijaart1999least, QP2025}. The dataset is available in \cite{mooijaart1999least, QP2025} as well as in the supplementary materials. The columns of the dataset have nearly identical sums. Before applying MVC-NMF and MAV-NMF, we normalize the dataset so that the sum of each column equals 1.

Table~\ref{T: resuoftimebudgetk3} reports the results of MVC-NMF and MAV-NMF for $K = 3$, where M3580 denotes males aged 35-49 in the 1980 survey and F5075 denotes females aged 50-64 in the 1975 survey, and similarly for the other group labels. In addition, the table includes the results of $K = 1$ as reference, where $\bm{M}$ is the average of the row sums of the time-allocation dataset provided in the supplementary materials and $\bm{H}$ is a matrix of ones. From the table, we observe that MAV-NMF tends to produce a sparser basis matrix $\bm{M}$ and a less sparse coefficient matrix $\bm{H}$ compared with MVC-NMF. This sparsity in the basis matrix improves interpretability: by comparing $\bm{M}$ for $K = 1$, dimension~1 is mainly associated with domestic work and caring for household members, dimension~2 with paid work, and dimension~3 with education. Table~\ref{T: voltimebudget} reports $\log\det(\bm{M}^\top\bm{M} + \delta \bm{I})$, again showing that MAV-NMF yields a larger volume than MVC-NMF.

\begin{table}[H]
\tiny
\centering  
\caption{Time allocation dataset for dimensionality $K = 3$ using MVC-NMF and MAV-NMF, where M3580 denotes males aged 35-49 in the 1980 survey and F5075 denotes females aged 50-64 in the 1975 survey, and similarly for the other group labels.} 
\label{T: resuoftimebudgetk3}
    \centering
\begin{tabular}{l|l|lll|lll}
  \hline
\multirow{1}{*}{Model} &  \multicolumn{1}{c|}{/} &\multicolumn{3}{c|}{MVC-NMF}&
\multicolumn{3}{c}{MAV-NMF} \\
\cline{1-8}
$\bm{M}$& $k=1$ & $k=1$ & $k=2$ & $k=3$ & $k=1$ & $k=2$ & $k=3$\\
\hline
\text{paidwork} & 0.0803 & 0.0000 & 0.2180& 0.0630 & 0.0000 & 0.2499 & 0.0000\\
\text{dom.work} & 0.0779 & 0.1471 & 0.0171& 0.0042 & 0.1460 & 0.0000 & 0.0000\\
\text{caring} & 0.0167 & 0.0241 & 0.0162& 0.0000 & 0.0237 & 0.0133 & 0.0000\\
\text{shopping} & 0.0253 & 0.0360 & 0.0165& 0.0131 & 0.0360 & 0.0137 & 0.0104\\
\text{per.need} & 0.0347 & 0.0371 & 0.0318& 0.0332 & 0.0371 & 0.0310 & 0.0336\\
\text{eating} & 0.0623 & 0.0642 & 0.0675& 0.0496 & 0.0642 & 0.0683 & 0.0401\\
\text{sleeping} & 0.3581 & 0.3635 & 0.3348& 0.3811 & 0.3637 & 0.3301 & 0.4034\\
\text{educat.} & 0.0335 & 0.0000 & 0.0000& 0.1694 & 0.0000 & 0.0000 & 0.2571\\
\text{particip} & 0.0146 & 0.0156 & 0.0163& 0.0094 & 0.0156 & 0.0166 & 0.0056\\
\text{soc.cont} & 0.0656 & 0.0776 & 0.0587& 0.0473 & 0.0776 & 0.0561 & 0.0404\\
\text{goingout} & 0.0297 & 0.0220 & 0.0321& 0.0444 & 0.0220 & 0.0334 & 0.0515\\
\text{sports} & 0.0363 & 0.0426 & 0.0230& 0.0417 & 0.0427 & 0.0199 & 0.0505\\
\text{gardening} & 0.0194 & 0.0200 & 0.0258& 0.0083 & 0.0200 & 0.0267 & 0.0000\\
\text{outside} & 0.0061 & 0.0057 & 0.0075& 0.0049 & 0.0057 & 0.0079 & 0.0034\\
\text{tv-radio} & 0.0779 & 0.0773 & 0.0789& 0.0777 & 0.0775 & 0.0793 & 0.0758\\
\text{reading} & 0.0359 & 0.0419 & 0.0344& 0.0236 & 0.0420 & 0.0335 & 0.0170\\
\text{relaxing} & 0.0070 & 0.0076 & 0.0058& 0.0074 & 0.0076 & 0.0055 & 0.0081\\
\text{other} & 0.0188 & 0.0198 & 0.0154& 0.0216 & 0.0198 & 0.0147 & 0.0245\\
   \hline
$\bm{H}^T$& $k=1$ & $k=1$ & $k=2$ & $k=3$ & $k=1$ & $k=2$ & $k=3$\\
\hline
\text{M1275} & 1.0000 & 0.0000 & 0.1623& 0.8377 & 0.0805 & 0.3662 & 0.5534\\
\text{M1280} & 1.0000 & 0.0561 & 0.1005& 0.8434 & 0.1304 & 0.3137 & 0.5559\\
\text{M1285} & 1.0000 & 0.0480 & 0.0593& 0.8927 & 0.1205 & 0.2913 & 0.5882\\
\text{M2575} & 1.0000 & 0.0000 & 0.9542& 0.0458 & 0.1232 & 0.8384 & 0.0384\\
\text{M2580} & 1.0000 & 0.0349 & 0.8541& 0.1109 & 0.1504 & 0.7691 & 0.0805\\
\text{M2585} & 1.0000 & 0.0805 & 0.8192& 0.1003 & 0.1908 & 0.7362 & 0.0730\\
\text{M3575} & 1.0000 & 0.0897 & 0.8594& 0.0509 & 0.2037 & 0.7582 & 0.0381\\
\text{M3580} & 1.0000 & 0.0682 & 0.8899& 0.0419 & 0.1844 & 0.7820 & 0.0336\\
\text{M3585} & 1.0000 & 0.0492 & 0.9256& 0.0252 & 0.1678 & 0.8084 & 0.0238\\
\text{M5075} & 1.0000 & 0.1542 & 0.7876& 0.0582 & 0.2612 & 0.6984 & 0.0404\\
\text{M5080} & 1.0000 & 0.2721 & 0.6337& 0.0943 & 0.3624 & 0.5751 & 0.0625\\
\text{M5085} & 1.0000 & 0.3407 & 0.5709& 0.0885 & 0.4225 & 0.5196 & 0.0579\\
\text{M6575} & 1.0000 & 0.6952 & 0.1508& 0.1539 & 0.7309 & 0.1740 & 0.0951\\
\text{M6580} & 1.0000 & 0.6956 & 0.0989& 0.2055 & 0.7299 & 0.1433 & 0.1268\\
\text{M6585} & 1.0000 & 0.7017 & 0.1085& 0.1897 & 0.7351 & 0.1475 & 0.1174\\
\text{F1275} & 1.0000 & 0.2410 & 0.1168& 0.6422 & 0.3004 & 0.2738 & 0.4258\\
\text{F1280} & 1.0000 & 0.2076 & 0.0588& 0.7336 & 0.2662 & 0.2481 & 0.4857\\
\text{F1285} & 1.0000 & 0.1886 & 0.0000& 0.8114 & 0.2463 & 0.2173 & 0.5363\\
\text{F2575} & 1.0000 & 0.8230 & 0.1770& 0.0000 & 0.8431 & 0.1547 & 0.0022\\
\text{F2580} & 1.0000 & 0.8126 & 0.1845& 0.0028 & 0.8291 & 0.1603 & 0.0105\\
\text{F2585} & 1.0000 & 0.6889 & 0.2964& 0.0147 & 0.7214 & 0.2602 & 0.0184\\
\text{F3575} & 1.0000 & 0.9102 & 0.0898& 0.0000 & 0.9192 & 0.0808 & 0.0000\\
\text{F3580} & 1.0000 & 0.8945 & 0.1055& 0.0000 & 0.9055 & 0.0945 & 0.0000\\
\text{F3585} & 1.0000 & 0.8632 & 0.1368& 0.0000 & 0.8785 & 0.1215 & 0.0000\\
\text{F5075} & 1.0000 & 0.9690 & 0.0310& 0.0000 & 0.9699 & 0.0301 & 0.0000\\
\text{F5080} & 1.0000 & 0.9672 & 0.0328& 0.0000 & 0.9686 & 0.0314 & 0.0000\\
\text{F5085} & 1.0000 & 0.9192 & 0.0806& 0.0002 & 0.9259 & 0.0725 & 0.0016\\
\text{F6575} & 1.0000 & 0.9501 & 0.0049& 0.0451 & 0.9534 & 0.0190 & 0.0275\\
\text{F6580} & 1.0000 & 0.9683 & 0.0000& 0.0317 & 0.9693 & 0.0096 & 0.0211\\
\text{F6585} & 1.0000 & 0.9342 & 0.0088& 0.0570 & 0.9391 & 0.0259 & 0.0350\\
   \hline
\end{tabular}
\end{table}

\begin{table}[H]
\caption{Real-world time allocation dataset; $\text{logdet}(\bm{M}^T\bm{M} + \delta \bm{I})$}
    \label{T: voltimebudget}
    \centering
    \begin{tabular}{c|c}
    \hline 
    Methods & Time allocation\\
    \hline
 MVC-NMF &  -4.6060  \\
  MAV-NMF & -4.2750 \\
    \hline
    \end{tabular}
\end{table} 

\section{Conclusion}\label{S: Conclusion}

This paper introduces a new NMF framework, termed maximum-volume-constrained NMF (MAV-NMF), which encourages the learned basis vectors to be as distinct as possible. In contrast to minimum-volume-constrained NMF (MVC-NMF), MAV-NMF is particularly suitable for highly mixed data, where the rows of the coefficient matrix do not contain zero entries, and it tends to yield more interpretable and unmixed basis vectors. We provide a sufficient condition, related to the sufficiently scattered condition, under which the solution of MAV-NMF is unique. To estimate MAV-NMF, we exploit an equivalent transformation that replaces the negative determinant regularization on the basis matrix with the positive one on
the coefficient matrix. Finally, we demonstrate the performance of MAV-NMF on two artificial sedimentary geology datasets, three additional datasets generated by us, a human face images dataset, and a social science time-allocation dataset.

\section*{Acknowledgments}

Zhongming Chen is partially supported by Natural Science Foundation of Zhejiang Province (No. LY22A010012) and Natural Science Foundation of Xinjiang Uygur Autonomous Region (No. 2024D01A09).

\section*{Competing Interests}

No potential competing interest was reported by the authors.

\newpage
\bibliographystyle{unsrt}
\bibliography{references.bib}

\appendixqq
\appendixqqsection{Proof of Theorem 1}
The proof of Theorem~\ref{theorem: maxvol} follows arguments similar to those in \cite{fu2015, leplat2020, fu2018identifiability} and proceeds as follows.

Step 1: Assume that both ($\tilde{\bm{M}}, \tilde{\bm{H}}$) and ($\bm{M}_{\#}$, $\bm{H}_{\#}$) are optimal solutions for (\ref{E: maxvolcriteria}). Since 
\begin{equation*}
    \text{rank}(\tilde{\bm{M}}) = \text{rank}(\tilde{\bm{H}}) = \text{rank}(\bm{M}_{\#}) = \text{rank}(\bm{H}_{\#}) = K,
\end{equation*}
there exists a full rank matrix $\bm{S}\in \Re_+^{K\times K}$ such that 
\begin{equation*}
    \bm{M}_{\#} = \tilde{\bm{M}}\bm{S} \text{ and } \bm{H}_{\#} = \bm{S}^{-1}\tilde{\bm{H}}.
\end{equation*}
Moreover, $\bm{S}^{-1}$ satisfies the column-sum-to-1 constraint since
\begin{equation*}\label{eq: sumto1}
    \bm{1}^T\bm{S}^{-1} = \bm{1}^T\bm{H}_{\#}\tilde{\bm{H}}^\dag = \bm{1}^T\tilde{\bm{H}}^\dag = (\bm{1}^T\tilde{\bm{H}})\tilde{\bm{H}}^\dag = \bm{1}^T,
\end{equation*}
where $\tilde{\bm{H}}^\dag$ denotes a right inverse of $\tilde{\bm{H}}$, i.e., $\tilde{\bm{H}}\tilde{\bm{H}}^\dag = \bm{I}_K$;
this inverse exists since $\text{rank}(\tilde{\bm{H}}) = K$. Consequently,
\begin{equation}\label{eq: usumto1}
    \bm{1}^T\bm{S} = \bm{1}^T
\end{equation}

Step 2: From $\bm{M}_{\#} = \tilde{\bm{M}}\bm{S}$, each column of $\bm{S}$ belongs to the dual cone of $\tilde{\bm{M}}^T$, i.e.,
\begin{equation*}
    \bm{S}(:,k) \in \text{cone}^*(\tilde{\bm{M}}^T), \quad k = 1, \cdots, K.
\end{equation*}
By SSC1 ($\mathbb{C} \subseteq \text{cone}(\tilde{\bm{M}}^T)$), it follows that $\text{cone}^*(\tilde{\bm{M}}^T) \subseteq \mathbb{C}^*$. Hence, $\bm{S}(:, k) \in \mathbb{C}^*$. Therefore,
\begin{equation}\label{eq: dualc}
    ||\bm{S}(:, k)||_2 \leq \bm{1}^T\bm{S}(:, k), \quad k = 1, \cdots, K.
\end{equation}
Consequently,
\begin{equation*}
    |\text{det}(\bm{S})| \leq \prod_k||\bm{S}(:, k)||_2 \leq \prod_k \bm{1}^T\bm{S}(:, k) = 1.
\end{equation*}
where the first inequality follows from Hadamard's inequality, the second from (\ref{eq: dualc}), the last equality from (\ref{eq: usumto1}).

Step 3:  If $|\text{det}(\bm{S})| < 1$, then 
\begin{equation*}
\begin{split}
    \text{det}(\bm{M}_{\#}^T\bm{M}_{\#}) 
    & = \text{det}(\bm{S}^T\tilde{\bm{M}}^T\tilde{\bm{M}}\bm{S}) 
    \\&= |\text{det}(\bm{S})|^2\text{det}(\tilde{\bm{M}}^T\tilde{\bm{M}}) 
    \\&< \text{det}(\tilde{\bm{M}}^T\tilde{\bm{M}}).
\end{split}
\end{equation*}
which contradicts the optimality of $(\bm{M}_{\#}, \bm{H}_{\#})$ for the maximum-volume criteria.

Step 4: If $|\text{det}(\bm{S})| = 1$, then all the above inequalities must hold with equality. Hence,   
\begin{equation*}
  ||\bm{S}(:, k)||_2 = \bm{1}^T\bm{S}(:, k), \quad k = 1, \cdots, K,
\end{equation*}
which implies that $\bm{S}(:, k)$ lies on the boundary of $\mathbb{C}^*$. Together with $\bm{S}(:, k) \in \text{cone}^*(\tilde{\bm{M}}^T)$ and SSC2, this yields 
\begin{equation*}
    \bm{S}(:, k) = \alpha_k \bm{e}_k.
\end{equation*}

Step 5: From the constraint $\bm{1}^T\bm{S} = \bm{1}^T$ in Equation~(\ref{eq: usumto1}), it follows that $\alpha_k = 1$ for all $k$. Therefore, $\bm{S}$ must be a permutation matrix.

As in minimum-volume-constrained NMF, the constraint $\bm{H}^T\bm{1} = \bm{1}$ in \eqref{E: maxvolcriteria} can be replaced by $\bm{H}\bm{1} = \bm{1}$ or $\bm{M}^T\bm{1} = \bm{1}$. The proof follows arguments similar to those developed for MVC-NMF in \cite{fu2015, leplat2020, fu2018identifiability} and Theorem~\ref{theorem: maxvol}. 

\appendixqqsection{Three artificial datasets}

We generate data $\bm{X} \in \mathbb{R}_+^{9\times 500}$ according to $\bm{X} = \bm{M}\bm{H}$, where the basis matrix $\bm{M}\in \mathbb{R}_+^{9\times 3}$ is nonnegative and the coefficient matrix $\bm{H}\in \mathbb{R}_+^{3\times 500}$ is nonnegative with each column summing to one. Specifically, the entries of submatrix formed by the first three rows of $\bm{M}$ are drawn independently from a uniform distribution between 0 and one, while the submatrix formed by the remaining six rows satisfies the sufficiently scattered condition (SSC) \cite{fu2015, fu2016robust, lin2015identifiability}.

\begin{figure}
\centering
 \begin{subfigure}[b]{0.32\linewidth}
\includegraphics[width=1\textwidth]{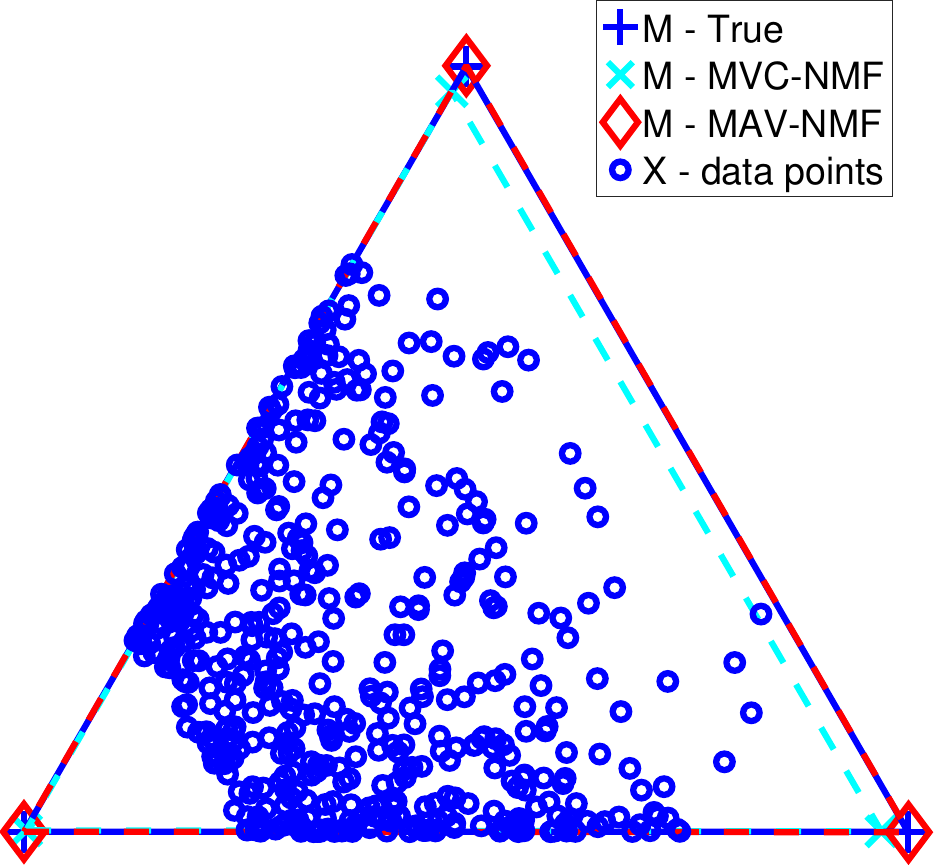}
 \caption{\footnotesize{One row without zero entries}}\label{F: minnomaxdirechlettwo}
 \end{subfigure}
 \begin{subfigure}[b]{0.32\linewidth}
\includegraphics[width=1\textwidth]{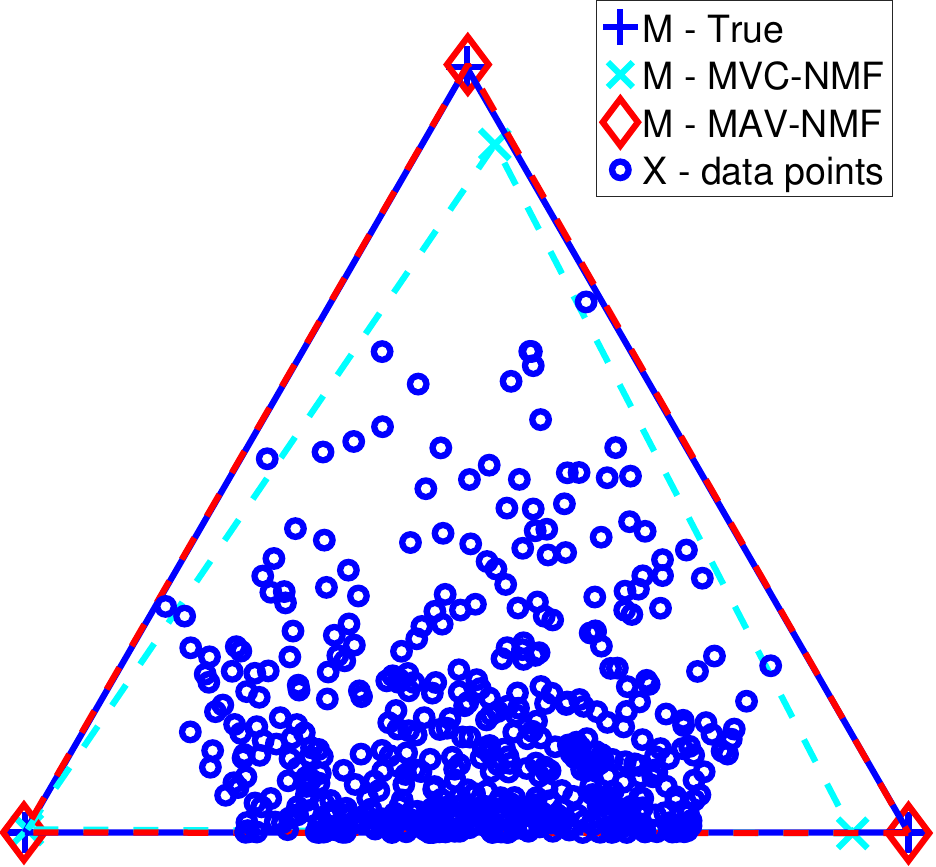}
 \caption{\footnotesize{Two rows without zero entries}}\label{F: minnomaxdirechletone}
 \end{subfigure}
 \begin{subfigure}[b]{0.32\linewidth}
\includegraphics[width=1\textwidth]{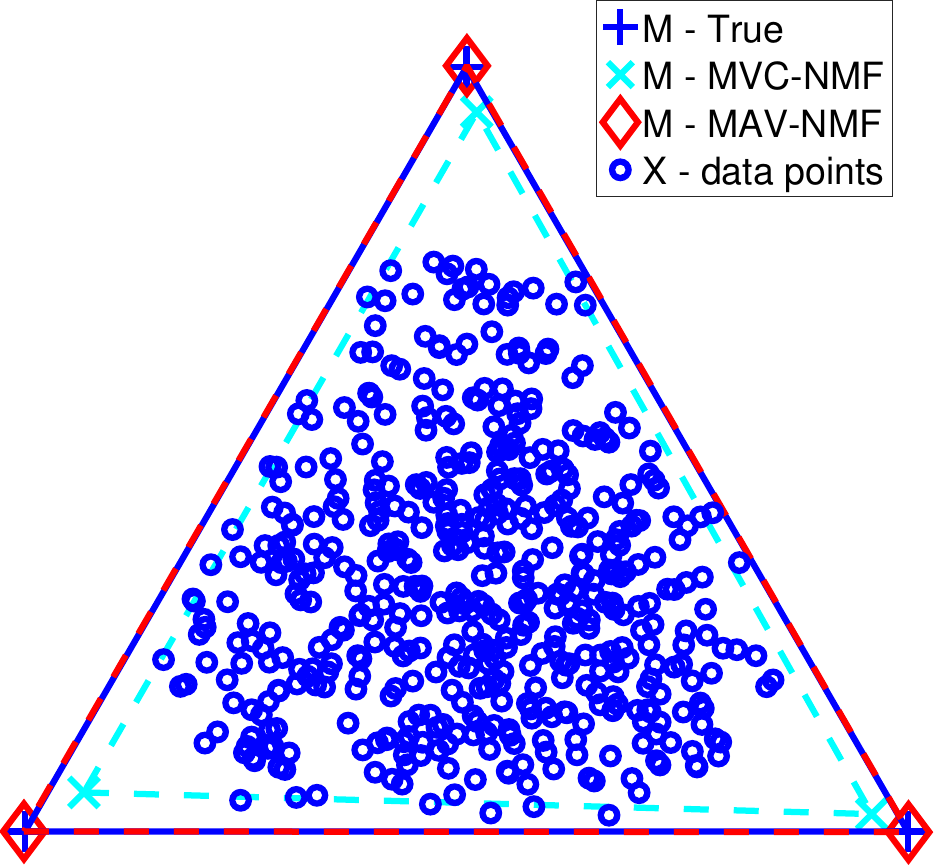}
 \caption{\footnotesize{Three rows without zero entries}}\label{F: minnomaxdirechletzero}
 \end{subfigure}
 \caption{Data with no zero entries in all rows of the ground truth coefficient matrix $\bm{H}$: (a) one row without zero entries; (b) two rows without zero entries; (c) all three rows without zero entries.}\label{F: direchlet}
    \end{figure}

We consider three data generation settings with different sampling schemes for $\bm{H}$.
In the first setting, each column of $\bm{H}$ is sampled from a Dirichlet distribution with parameter $(2, 0.5, 0.5)$ and resampled until all entries are not larger than $0.75$. Under this setting, one row of $\bm{H}$ contains no zero entries. Geometrically, one boundary of the simplex formed by the ground truth basis vectors does not contain any data points. As shown in Fig.~\ref{F: minnomaxdirechlettwo}, the blue circles denote data points, the blue plus signs indicate the ground truth basis vectors, and the solid blue triangle represents the simplex formed by the ground truth basis vectors. Note that in the figure, we take the ground truth basis vectors $\bm{M}(:, 1), \bm{M}(:, 2), \bm{M}(:, 3)$ as standard basis vectors $\bm{e}_1, \bm{e}_2, \bm{e}_3$. In the second setting, each column of $\bm{H}$ is sampled from a Dirichlet distribution with parameter $(2, 2, 0.5)$ and resampled until all entries are not larger than $0.75$, leading to two rows of $\bm{H}$ without zero entries; see Figure~\ref{F: minnomaxdirechletone}. In the third setting, each column of $\bm{H}$ is sampled from a Dirichlet distribution with parameter $(2, 2, 2)$ and resampled until all entries are not larger than $0.75$, so that none of the rows of $\bm{H}$ contains zero entries; see Figure~\ref{F: minnomaxdirechletzero}.

We apply minimum-volume-constrained NMF (MVC-NMF) and maximum-volume-constrained NMF (MAV-NMF) to the artificial data $\bm{X}$ to recover $\bm{M}$ and $\bm{H}$. Fig.~\ref{F: minnomaxdirechlettwo}, Fig.~\ref{F: minnomaxdirechletone}, and Fig.~\ref{F: minnomaxdirechletzero} show the results for the first, second, and third settings, respectively. In the figures, the cyan crosses and dashed cyan triangle correspond to the estimates obtained by MVC-NMF and the red diamonds and dashed red triangle correspond to MAV-NMF. We observe that MVC-NMF fails to recover the ground truth basis vectors, as MVC-NMF favors simplices with minimum volume, whereas the simplex formed by the ground truth basis vectors has a larger volume in these settings. In contrast, MAV-NMF achieves accurate recovery, since it can effectively expand the volume of the simplex.

Table~\ref{T: purityvol} reports the volume of the basis matrix, measured by $\text{logdet}(\bm{M}^T\bm{M} + \delta \bm{I})$. We observe that MAV-NMF has larger volume than MVC-NMF. For example, for data where only one edge of the simplex has data points (the third column in Table~\ref{T: purityvol}), the volumes for MVC-NMF and MAV-NMF are 1.1760 and 1.5040, respectively.
 
\begin{table}\scriptsize
\caption{The volume of basis matrix for data where not all rows of $\bm{H}$ contains zero entries; $\text{logdet}(\bm{M}^T\bm{M} + \delta \bm{I})$}
    \label{T: purityvol}
    \centering
    \begin{tabular}{c|c|c|c}
    \hline 
    Methods & one row without zero entries & two rows without zero entries & three rows without zero entries\\
    \hline
 MVC-NMF &  1.3720 & 1.1760 & 1.1030 \\
  MAV-NMF & 1.4990 & 1.5040 & 1.5030\\
    \hline
    \end{tabular}
\end{table}

\appendixqqsection{Social science: Time allocation dataset}

\begin{table}[H]
\centering  
\caption{Time-allocation dataset \cite{mooijaart1999least, QP2025}. 
} 
\label{T: timebudget}
    \centering
\rotatebox{90}
{
\resizebox{1.45\textwidth}{!}
{
\scriptsize
\setlength{\tabcolsep}{2pt}
\renewcommand{\arraystretch}{0.85}
\begin{tabular}{lrrrrrrrrrrrrrrrrrrrrrrrrrrrrrrr}
  \hline 
\multirow{3}{*}{Activities}&\multicolumn{15}{c}{\textbf{Male}} & \multicolumn{15}{c}{\textbf{Female}} \\
& \multicolumn{3}{c}{12--24} & \multicolumn{3}{c}{25--34} & \multicolumn{3}{c}{35--49} & \multicolumn{3}{c}{50--64} & \multicolumn{3}{c}{65+} &
\multicolumn{3}{c}{12--24} & \multicolumn{3}{c}{25--34} & \multicolumn{3}{c}{35--49} & \multicolumn{3}{c}{50--64} & \multicolumn{3}{c}{65+} \\
& 1975 & 1980 & 1985 & 1975 & 1980 & 1985 & 1975 & 1980 & 1985 & 1975 & 1980 & 1985 & 1975 & 1980 & 1985 &
1975 & 1980 & 1985 & 1975 & 1980 & 1985 & 1975 & 1980 & 1985 & 1975 & 1980 & 1985 & 1975 & 1980 & 1985 \\
  \hline
\text{paidwork} & 901 & 769 & 707 & 2180 & 1992 & 1899 & 1901 & 2008 & 2093 & 1708 & 1357 & 1206 & 176 & 71 & 95 & 723 & 665 & 564 & 439 & 471 & 704 & 299 & 375 & 412 & 151 & 153 & 233 & 11 & 6 & 19\\
\text{dom.work} & 87 & 157 & 155 & 250 & 269 & 341 & 249 & 289 & 331 & 244 & 337 & 450 & 617 & 563 & 636 & 494 & 460 & 397 & 1342 & 1338 & 1147 & 1567 & 1605 & 1529 & 1600 & 1558 & 1487 & 1319 & 1409 & 1318\\
\text{caring} & 33 & 28 & 15 & 194 & 206 & 184 & 99 & 128 & 136 & 51 & 54 & 25 & 124 & 27 & 38 & 135 & 99 & 86 & 635 & 673 & 651 & 296 & 309 & 308 & 83 & 84 & 82 & 78 & 154 & 44\\
\text{shopping} & 120 & 138 & 127 & 152 & 157 & 183 & 173 & 157 & 185 & 227 & 221 & 230 & 273 & 251 & 264 & 208 & 200 & 223 & 347 & 336 & 336 & 372 & 347 & 373 & 376 & 335 & 352 & 384 & 292 & 320\\
\text{per.need} & 289 & 294 & 316 & 293 & 316 & 302 & 351 & 339 & 332 & 350 & 364 & 352 & 365 & 392 & 383 & 359 & 377 & 387 & 311 & 339 & 337 & 325 & 346 & 351 & 367 & 368 & 385 & 372 & 453 & 366\\
\text{eating} & 508 & 528 & 527 & 623 & 649 & 605 & 660 & 709 & 650 & 709 & 744 & 686 & 763 & 767 & 707 & 536 & 513 & 495 & 593 & 607 & 572 & 664 & 633 & 656 & 601 & 613 & 595 & 635 & 665 & 615\\
\text{sleeping} & 3737 & 3765 & 3744 & 3380 & 3403 & 3397 & 3463 & 3445 & 3347 & 3560 & 3569 & 3533 & 3801 & 3871 & 3694 & 3744 & 3777 & 3821 & 3526 & 3532 & 3447 & 3567 & 3554 & 3444 & 3673 & 3701 & 3566 & 3849 & 3713 & 3675\\
\text{educat.} & 1447 & 1455 & 1537 & 124 & 245 & 208 & 56 & 90 & 85 & 18 & 58 & 46 & 10 & 43 & 54 & 1163 & 1321 & 1436 & 77 & 115 & 120 & 104 & 98 & 68 & 27 & 30 & 40 & 6 & 21 & 23\\
\text{particip} & 128 & 101 & 92 & 129 & 126 & 143 & 195 & 156 & 148 & 122 & 207 & 272 & 159 & 192 & 214 & 125 & 88 & 80 & 85 & 115 & 115 & 133 & 143 & 196 & 195 & 179 & 195 & 108 & 124 & 139\\
\text{soc.cont} & 515 & 505 & 449 & 609 & 649 & 599 & 671 & 593 & 479 & 603 & 704 & 554 & 811 & 671 & 619 & 592 & 557 & 527 & 780 & 776 & 736 & 694 & 689 & 699 & 758 & 810 & 721 & 929 & 796 & 749\\
\text{goingout} & 490 & 396 & 441 & 382 & 321 & 391 & 360 & 240 & 336 & 237 & 279 & 264 & 213 & 220 & 274 & 364 & 400 & 396 & 316 & 270 & 303 & 225 & 229 & 277 & 255 & 212 & 268 & 219 & 187 & 202\\
\text{sports} & 419 & 436 & 485 & 269 & 279 & 271 & 206 & 280 & 291 & 209 & 299 & 316 & 297 & 403 & 476 & 348 & 370 & 352 & 306 & 352 & 368 & 335 & 440 & 453 & 323 & 504 & 545 & 297 & 482 & 579\\
\text{gardening} & 111 & 102 & 100 & 173 & 213 & 231 & 259 & 238 & 268 & 256 & 288 & 309 & 366 & 312 & 308 & 90 & 76 & 86 & 149 & 131 & 145 & 198 & 154 & 170 & 197 & 190 & 217 & 169 & 191 & 169\\
\text{outside} & 48 & 41 & 64 & 69 & 35 & 67 & 88 & 45 & 64 & 116 & 76 & 112 & 86 & 117 & 178 & 32 & 32 & 41 & 41 & 32 & 44 & 37 & 34 & 45 & 53 & 41 & 76 & 37 & 39 & 52\\
\text{tv-radio} & 752 & 815 & 860 & 700 & 671 & 812 & 785 & 804 & 812 & 921 & 862 & 1012 & 1161 & 1198 & 1233 & 594 & 581 & 702 & 547 & 497 & 565 & 622 & 576 & 582 & 710 & 644 & 708 & 888 & 860 & 1076\\
\text{reading} & 272 & 256 & 188 & 366 & 318 & 243 & 316 & 343 & 319 & 468 & 413 & 467 & 477 & 660 & 578 & 292 & 257 & 207 & 300 & 275 & 265 & 356 & 311 & 307 & 478 & 390 & 377 & 485 & 404 & 460\\
\text{relaxing} & 78 & 56 & 73 & 64 & 58 & 57 & 59 & 44 & 58 & 79 & 57 & 68 & 157 & 92 & 104 & 73 & 74 & 63 & 88 & 54 & 63 & 76 & 63 & 59 & 78 & 53 & 64 & 63 & 67 & 69\\
\text{other} & 146 & 240 & 200 & 124 & 172 & 148 & 188 & 170 & 146 & 203 & 190 & 174 & 223 & 230 & 225 & 208 & 234 & 214 & 199 & 167 & 164 & 207 & 174 & 153 & 154 & 212 & 170 & 230 & 216 & 204\\
   \hline
\end{tabular}}}
\end{table}

\end{document}